\documentclass{article}

\usepackage{microtype}
\usepackage{graphicx}
\usepackage{subfigure}
\usepackage{booktabs} % for professional tables

\usepackage{pbalance}
\usepackage[notransparent]{svg}

\usepackage{hyperref}

\usepackage[noend]{algorithmic}
\usepackage{xcolor}

 \usepackage[accepted]{icml2024}

\usepackage{amsmath}
\usepackage{amssymb}
\usepackage{mathtools}
\usepackage{amsthm}

\usepackage[capitalize,noabbrev]{cleveref}

\theoremstyle{plain}
\newtheorem{theorem}{Theorem}[section]

\theoremstyle{definition}

\theoremstyle{remark}

\usepackage[textsize=tiny]{todonotes}

\icmltitlerunning{Oracle-Checker Scheme for Evaluating a LLM}

\begin{document}

\twocolumn[
\icmltitle{Oracle-Checker Scheme for Evaluating a Generative Large Language Model}



\icmlsetsymbol{equal}{*}

\begin{icmlauthorlist}
	\icmlauthor{Yueling (Jenny) Zeng}{ucsb}
	\icmlauthor{Li-C. Wang}{ucsb}
	\icmlauthor{Thomas Ibbetson}{ucsb}
%\icmlauthor{Firstname1 Lastname1}{equal,yyy}
%\icmlauthor{Firstname2 Lastname2}{equal,yyy,comp}
%\icmlauthor{Firstname3 Lastname3}{comp}
%\icmlauthor{Firstname4 Lastname4}{sch}
%\icmlauthor{Firstname5 Lastname5}{yyy}
%\icmlauthor{Firstname6 Lastname6}{sch,yyy,comp}
%\icmlauthor{Firstname7 Lastname7}{comp}
%%\icmlauthor{}{sch}
%\icmlauthor{Firstname8 Lastname8}{sch}
%\icmlauthor{Firstname8 Lastname8}{yyy,comp}
%%\icmlauthor{}{sch}
%%\icmlauthor{}{sch}
\end{icmlauthorlist}

\icmlaffiliation{ucsb}{Department of Electrical and Computer Engineering, University of California, Santa Barbara, USA}
%\icmlaffiliation{yyy}{Department of XXX, University of YYY, Location, Country}
%\icmlaffiliation{comp}{Company Name, Location, Country}
%\icmlaffiliation{sch}{School of ZZZ, Institute of WWW, Location, Country}

\icmlcorrespondingauthor{Yueling Zeng}{yuelingzeng@ucsb.edu}
\icmlcorrespondingauthor{Li-C. Wang}{licwang@ucsb.edu}

\icmlkeywords{Machine Learning, ICML}

\vskip 0.3in
]



\printAffiliationsAndNotice{\icmlEqualContribution} % otherwise use the standard text.

\begin{abstract}
This work presents a novel approach called oracle-checker scheme for evaluating
the answer given by 
a generative large language model (LLM). Two types of checkers are presented. 
The first type of checker follows the idea of property testing. 
The second type of checker follows the idea of program checking. 
Their applications are demonstrated in two separate contexts, entity extraction
and paraphrase decision, respectively. 
\end{abstract}

\section{Introduction}
\label{Introduction}

Suppose one desires to compute a function $f$. 
The person has access to an {\it oracle} that claims to be able
to compute $f$. For an input $x$, the person asks the oracle
for an output. The oracle responds with a $y$. 
The person needs to decide whether or not to accept $y$. 
To do so, the person runs a {\it checker} designed specifically
for $f$. If the checker passes on the pair $(x,y)$, $y$
is accepted and otherwise, it is rejected. 

Intuitively the checker needs to provide a definition of $f$
because the oracle's definition of $f$ might not be the
person's definition.  
Then, under the checker's definition, the result $y=f(x)$ is checked. 
For an $f$, if an oracle-checker pair exists 
we call it an {\it oracle-checker} scheme for computing $f$.

In this work, we study if a generative large language model (LLM)
can be used as an oracle. 
We consider {\it entity extraction} and {\it paraphrase decision} as
two underlying functions to be performed. The focus is on the
development of the corresponding checkers.
The objective is to show the fundamental ideas for 
realizing an oracle-checker scheme. 

The realization of an oracle-checker scheme in view of treating
a generative LLM as an oracle, is useful. 
It is common that the output of a generative LLM can vary 
on repeated runs of a fixed prompt. It is also common that 
an answer provided by a generative LLM requires validation. 

Automatically validating an answer given by a LLM can be an  
intriguing problem. This is especially the case when using
a labeled dataset for defining the
function $f$ is not sufficient. For example, one desires
to use a LLM for entity extraction. 
However, the person has a subjective view
of what an ``entity'' means, leading to a subjective view
on trustworthiness.
%The person might not even be able to
%define this view in advance, and only when the person
%sees an LLM's response, its 
%acceptance can be decided then. 
This view might be reflected in the person's experience 
with the LLM, where the person tries to decide the acceptance 
of an LLM's response after seeing them. 

\begin{figure*}[htb]
	\centering
	\includesvg[inkscapelatex=false, width=6.5in]{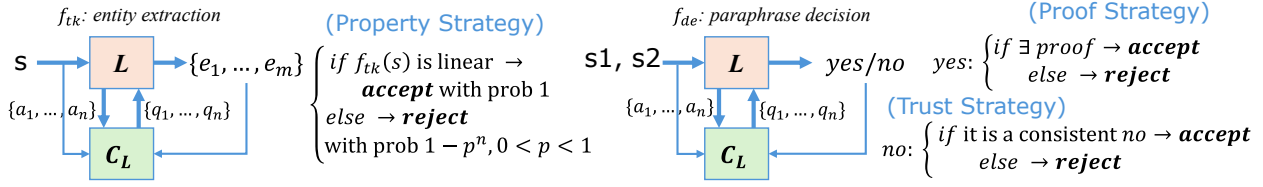}
	\caption{For a large LLM $L$, a checker $C_L$ can follow one of the three strategies. 
	For entity extraction, $C_L$ implements the property strategy. 
For paraphrase decision, $C_L$ implements the proof strategy if the answer is yes,
and the trust strategy if the answer is no.}
	\label{overview}
\end{figure*}

In this sense, the job of a checker is to articulate the person's
subjective view. In this work, we show that there can be three 
strategies for such articulation. 

Under the {\it property} strategy, the person believes that
the computation of $f$ should satisfy a certain property. 
For example, $f$ should follow a certain form of linear complexity. 
Then, the checker verifies if a given $(x,y)$ satisfies this
assumption. In this work, we show how to design a linear-complexity
checker for entity extraction. 

Under the {\it proof} strategy, the person accepts an answer
if a {\it proof} can be constructed.
Constructing a proof may require further interactions with 
the oracle. 
Then, under the {\it trust} strategy, the person accepts an answer
if the oracle passes a type of truthfulness test. In this case, the person
has no idea how the oracle computes $f$, 
or having any proof on the correctness of the answer. 
The answer is accepted because the person believes
the oracle's answer is consistent in view of the test and hence, 
is likely being truthful. By using {\it paraphrase decision} as an example, 
we show how to design these two types of checkers. 

{\bf Main contributions} 
of the work include the following three aspects:
(1) on the novel ideas for realization of the proposed oracle-checker scheme,
(2) on demonstrating how the scheme can be implemented in two 
separate contexts: entity extraction and paraphrase decision, and
(3) on the application of different checkers, showing how they can help
	assess the trustworthiness of a generative LLM. 

For the rest of the paper, 
Section~\ref{sec02} introduces the oracle-checker scheme and
its background. 
Section~\ref{sec03} presents a linearity
checker for entity extraction.
Section~\ref{sec04} presents our checker designs
for paraphrase decision. 
Sections~\ref{sec05} explains 
their implementation and discusses experiment results.
Section~\ref{sec06} reviews related works and 
section~\ref{sec07} concludes.

\section{The oracle-checker scheme}
\label{sec02}

In the oracle-checker scheme, an oracle exists
because of the checker, and vice versa. Given a function
$f$, a LLM $L$ is an oracle if a corresponding
checker $C_L$ can be realized. 

On an input $x$, $L$ provides the output $y$ as an answer
for $f(x)$. $C_L$ checks for the acceptance of $y=f(x)$. 
$C_L$ prepares a series of queries
$q_1, \ldots, q_n$ to $L$. $L$ responds with additional
answers $a_1, \ldots, a_n$, respectively. 
Based on those answers, $C_L$ decides to
accept $y=f(x)$ or not. 

We classify $f$ into two types. An $f$ can be a task,
asking the oracle to generate a result. An $f$ can be
a decision function, asking a yes/no question. 
We use $f_{tk}$ to denote the first type and 
$f_{de}$ to denote the second type.

For an $f_{tk}$, the checker run a test $T_{tk}$ that
checks if $f_{tk}$ meets a certain property. If 
$f_{tk}$ indeed has the property, the checker accepts
the result. Otherwise, the checker reject
$f_{tk}$ with a high probability. 
See Figure~\ref{overview} for its overview. 

For an $f_{de}$, there are {\it two} cases. When the oracle's
answer is yes, the checker tries to construct a proof. 
If a proof can be constructed, the
checker accepts the answer. Otherwise, it rejects. 
When the answer is no, the checker follows the 
trust strategy by verifying that the answer is truthful.

\subsection{Theoretical background}
\label{sec02.1}

Our oracle-checker scheme follows the ideas of
{\it program checking} pioneered by \cite{BlumKanna95}\cite{BlumLubyRubinfeld90}
in the 90's. 
Those ideas later inspired the development of the {\it property testing} field
\cite{Dana10}\cite{Goldreich17}, which had been studied for decades. 
Our work is based on two of the earliest ideas: the linearity test  
\cite{BlumLubyRubinfeld90} and the checking of graph isomorphism
\cite{BlumKanna95}. The first idea translates into our property strategy.
The second idea translates into our proof and trust strategies. 

\section{A linearity test for entity extraction}
\label{sec03}

Given two groups $G_1$ and $G_2$, let $f$ be a function
$f: G_1 \rightarrow G_2$. $f$ is called a group homomorphism if
$\forall x,z \in G_1$, it holds that $f(x+z)=f(x)+f(z)$. 
Testing group homomorphism is also referred as linearity test. 

The test is simple: Uniformly sample $x,z$ from $G_1$ $n$ times
and check if $f(x+z)=f(x)+f(z)$ holds. The $n$ is a variable depending
on what level of confidence to state about the linearity of $f$
after all $n$ tests pass. 

In property testing, a property of $f$ is measured as a distance 
between two functions. The distance between $f$ and $g$,
denoted as $\delta(f,g)$, is given as $Pr_{x}[f(x)=g(x)]$ where
the random $x$ is drawn uniformly from the domain. Let ${\cal F}$
be a family of functions that have the same domain as $f$ and
have the property (e.g. linearity). The distance from $f$ to ${\cal F}$,
$\delta(f,{\cal F})$, is defined as $\mbox{min}_{g \in {\cal F}}\{\delta(f,g)\}$. 
$f$ is said to be $\epsilon$-close to ${\cal F}$ if 
$\delta(f,{\cal F}) \leq \epsilon$ and otherwise, $f$ is $\epsilon$-far. 

\begin{theorem}
	\label{linearity1}
	Each linearity test provides its own assurance:
	If $f$ is a homomorphism from $G_1$ to $G_2$, the test passes
	with probability 1. If $f$ is $\epsilon$-far from ${\cal F}$ which
	is the set of all homomorphism from $G_1$ to $G_2$, then the 
	test fails with a probability at least $3\epsilon - 6 \epsilon^2$
	when $\epsilon \leq \frac{1}{4}$ and at least $\frac{2}{9}$ when otherwise
	(in this case, the $p$ in Figure~\ref{overview} is $\frac{7}{9}$). 
\end{theorem}

\begin{theorem}
	\label{linearityn}
	Continuing from Theorem~\ref{linearity1},
	by running the linearity test $n$ times, 
	$n= \Theta(\frac{1}{\epsilon})$, the checker accepts
	with probability 1 if $f$ is a homomorphsim, and otherwise rejects
	with a probability at least $\frac{2}{3}$.   
\end{theorem}

Detail of the two theorems can be found in \cite{Goldreich17} and
\cite{Dana10}, respectively. Practically, the theorems provide 
a way to {\it bound} the meaning of an acceptance by the linearity checker. 
For example, if the checker runs 10 tests and they 
all pass, then the checker accepts. This acceptance means {\it at least}
that one can say $f$ is $0.1$-close ($\epsilon=0.1$) from homomorphism with probablilty 
roughly 0.936 ($\approx 1-(1-3\epsilon+6\epsilon^2)^{10}$) or is
$0.05$-close ($\epsilon=0.05$) with probability roughly 0.765.  
It is worth noting that the probabilities are theoretical bounds
and the two theorems do not mention the group sizes,
i.e. the bounds are independent of the number
of elements in the groups. 

\subsection{$G_1,G_2$ in entity extraction}
\label{sec03.1}

\begin{figure}[htb]
	\centering
	\includesvg[inkscapelatex=false, width=3.2in]{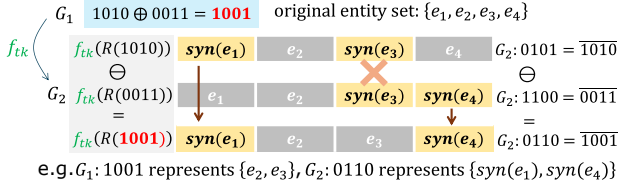}
	\caption{For linearity test in entity extraction, the domain group $G_1$
		is defined based on a set of binary vectors and the XOR ($\oplus$) operator.
		The range group $G_2$ is defined based on a set of synonymous entities
		(generated by the synonym generator $syn$) and the XNOR
		operator $\ominus$ on its subsets, also represented as vectors. }
	\label{homo}
\end{figure}

To realize the linearity test in the context of entity extraction,
we need to define the meaning of the two groups $G_1, G_2$.  
Let $L_{e}$ denote the prompt to a LLM $L$ for entity extraction.
Given a sentence $s$, $L_e(s) \rightarrow \{e_1, \ldots, e_m\}$,
a set of $m$ entities. 
The test is applied on $L_e(s)$ for each $s$ individually. 
We let $G_1$ be the powerset of  $L_e(s)$, denoted as $2^{L_e(s)}$. 

Let $syn$ be a {\it synonym generator} that for each $e_i$,
$syn(e_i)$ generates a synonym $e'_i$. 
Let $SE$ be the set of synonyms $\{e'_1, \ldots, e'_m\}$ by
applying $syn$ on  $L_e(s)$. 
We let $G_2$ be $2^{SE}$.

For each subset of  $2^{L_e(s)}$, it can be represented
as a binary characteristic vector $v=[b_1, \ldots, b_m]$. 
For a $b_i=0$, let it mean that $e_i$ is present in the subset
and otherwise, is not present.
We create a replacement function $R$ that 
given a vector $v$, if a $b_j=1$ we replace $e_j$ with
its synonym $e'_j$ in $s$, i.e. $R(v)$ produces a new
sentence $s_v$ with some entities replaced by their
synonyms according to $v$.

Similarly, we use a binary vector $u=[b'_1, \ldots, b'_m]$
to represent each element of $G_2$. 
Let $f_{tk}$ be the ideal entity extraction function. 
We define the mapping from $G_1$ to $G_2$ to be:
$f_{tk}(R(v)) \rightarrow \overline{v}$ where $\overline{v}$
is the complementary vector of $v$.   

The $G_1, G_2$ as defined are clearly groups. 
The set of each group includes all bit vector of length $m$.
In $G_1$, the operator is XOR ($\oplus$) and in
$G_2$, is XNOR ($\ominus$). Let $f(v)=f_{tk}(R(v))$.
The homomorphism from $G_1$ to $G_2$ is stated as:
$\forall x,y, f(x \oplus y)= f(x) \ominus f(y)$.
Figure~\ref{homo} illustrates this mapping with an
example of four entities. 

\subsection{Key points in the linearity test}
\label{sec03.2}

When one asks the LLM $L$ to perform entity extraction,
the meaning of ``entity'' is not always well defined. 
By applying the test, we are saying that only those
entities passing the test meet our definition of
so-called ``entities''. 

When $G_1$ to $G_2$ is indeed a homomorphism, 
for an extracted entity $e_i$, its extraction is not
sensitive to replacing $e_i$ with its synonym, nor is it
sensitive to replacing other entities. 
In a sense, the extraction of $e_i$ is independent
of the extraction of others. In contrast, suppose
$e_i$ and $e_j$ must be extracted ``together'', i.e. 
one cannot be extracted without the other. Then, the
homomorphism property is violated. 

Note that the test depends on two methods: 
the synonym generator $syn$ and the entity replacement
$R$. Hence, the practical meaning of the test depends
on how we implement the two. For example, $R$ can be
as simple as replacing the entities in the original sentence,
or to be more general, it can be based on an entity-persevered 
paraphrase. In the second case, the implementation can be more
complicated but the scope of the test 
is larger. The experimental properties of the test will be
discussed in detail in Sections~\ref{sec05.01}. 

\section{Program checking for paraphrase decision}
\label{sec04}

Given two graphs $g_1, g_2$ of $N$ vertices
labeled by $\{1,\ldots, N\}$. The two graphs is
isomorphic if there exists a permutation $\pi$ on the
labels such that $\pi$ transforms
$g_1$ into $g_2$, i.e. $\pi: g_1 \rightarrow g_2$.
This is one of the earliest problems under study
for program checking \cite{BlumKanna95}. 

Suppose an oracle claims that $g_1, g_2$ are
isomorphic. In theory, checking this ``yes'' answer
is straightforward by asking the oracle to provide
a $\pi$ that transforms $g_1$ into $g_2$. 
If the oracle is limited to only providing a yes/no answer and cannot
provide a $\pi$, one can search for a satisfying $\pi$ 
by interacting with the oracle \cite{BlumKanna95}. 
This $\pi$ serves as a {\it proof} that $g_1, g_2$ are
isomorphic.  

In contrast, when the oracle's answer is ``no'' that
the $g_1, g_2$ are not isomorphic, this means that
there does not exist any $\pi$ that can transform $g_1$ into $g_2$. 
Then, finding a proof is difficult because it
needs to be shown on all permutations. 
Therefore, checking for ``no'' requires a
different strategy. 

The strategy involves two random choices. The first
is to randomly choose between $g_1$ and $g_2$. Let
this result be $h$. The second is to randomly choose
a permutation $\pi_i$ and apply it to $h$. 
Let the resulting graph be denoted as $\pi(h)$. 
The checker then asks the oracle which of the $g_1, g_2$
the $\pi(h)$ is produced from (see Figure~\ref{nonisomorphism} below). 
If $g_1, g_2$ are indeed non-isomorphic, the oracle
will be able to answer a series of such questions
without a mistake. Otherwise, the oracle can only
guess the answer purely randomly and if 
$n$ such tests are run, the probability that the 
oracle correctly answers all $n$ 
questions is only $\frac{1}{2^n}$. 

\subsection{From graph isomorphism to semantic equivalence}
\label{sec04.1}

Checking semantic equivalence   
is one of the basic functions in natural language processing. 
Suppose we have an oracle claimed to be able to tell 
semantic equivalence. To apply the above
program checking ideas, intuitively we can think the two
sentences as two graphs and the semantic equivalence 
as the isomorphic relation. Accordingly,
we need to define what a ``proof'' means for 
semantic equivalence. 

\begin{figure}[thb]
	\centering
	\includesvg[inkscapelatex=false, width=3in]{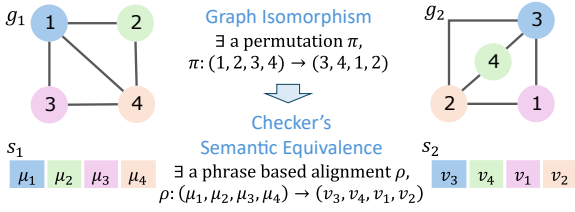}
	\caption{Mapping from graph isomorphism to semantic equivalence
	 by replace permutation $\pi$ with an alignment $\rho$ as the proof}
	\label{isomorphism}
	\vskip -0.1in
\end{figure}

In this work, we refer to the idea of
{\it compositional phrase alignment} (CPA)
\cite{arase-tsujii-2020-compositional}\cite{wang2010probabilistic}
as the basis for finding a proof for semantic equivalence.
Two sentences are said to be semantically equivalent if
an alignment $\rho$ (explained in detail later) 
can be established between them. 
Figure~\ref{isomorphism} illustrates this connection between
semantic equivalence and graph isomorphism. 

Let a consecutive sub-sequence of words in a sentence
be called a {\it phrase}. 
In Figure~\ref{isomorphism}, each $\mu_i$ and each
$\nu_j$ are phrases. 
The $\rho$ shows a {\it phrase mapping} from $s_1$ to $s_2$.
We call such a directional alignment as a $\rho$-{\it alignment},
to note its type.

With the alignment idea, when a LLM $L$ says that two sentences
are semantically equivalent, our
checker can try to find a $\rho$-alignment between them. 
However, when $L$'s answer is no, it is not as straightforward 
how to implement the graph non-isomorphism test accordingly. 
To follow the non-isomorphism test idea, 
the checker needs the capability that it can 
``pick a $\rho$-alignment randomly''
and use it to transform a random choice of $s_1$ and $s_2$. 
This implies that the checker needs to be very capable
in paraphrasing. 

\begin{figure}[htb]
	\centering
	\includesvg[inkscapelatex=false, width=3.2in]{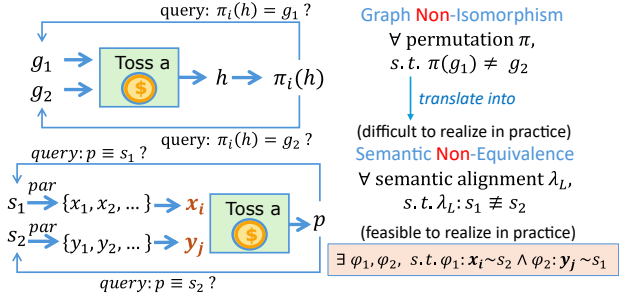}
	\caption{The scheme to approximate the non-isomorphism test} 	
	\label{nonisomorphism}
	%\vskip -0.1in
\end{figure}

It should be noted that a checker should be much simpler
than an oracle. Implementing an extensive 
paraphrasing capability inside
a checker for checking semantic equivalence, in a sense,
defeats the purpose of having a checker. In our checker,
we therefore rely on the oracle to provide random
paraphrases. Then, the checker will decide which
paraphrases can be used and which cannot. 

Figure~\ref{nonisomorphism} depicts our scheme to approximate the
graph non-isomorphism test in the context of semantic non-equivalence. 
The key is to observe that in the non-isomorphism context, the 
effectiveness of the test comes from oracle's inability to 
differentiate the source graph used to produce 
$\pi_i(h)$. This {\it indifferentiability} is what we should try 
to approximate. 

In Figure~\ref{nonisomorphism} we use ``$\equiv$'' to denote
semantic equivalence. When an oracle $L$ says that $s_1, s_2$
are not semantically equivalent, let us assume that $L$ has
checked that for all possible semantic alignment $\lambda_L$'s,
$s_1 \not\equiv s_2$. The checker has no idea what they are
and hence, cannot operate on them. 
Instead, the checker comes out with its own alignment scheme
$\phi$ which is based on finding a $\rho$-alignment.

Given $s_1, s_2$, the checker first asks the oracle to
randomly generate a set of paraphrases for each. 
In Figure~\ref{nonisomorphism}, the checker finds an $x_i$ (and $y_j$)
in the paraphrased set
of $s_1$ (and $s_2$) such that there exists a $\phi_1$ 
aligning $x_i$ to $s_2$ (and $\phi_2$ aligning $y_j$ to $s_1$). 
Checking $x_i$ being aligned to $s_2$,
is to enable the checker to say that
there can be a way to transform $s_2$ to $x_i$. 
Let us use ``$s_2 \rightsquigarrow x_i$'' to denote this. 

Recall that the oracle already claims that $s_1 \not\equiv s_2$.
Suppose $s_1 \equiv s_2$ and suppose our checker can also
find a $\phi_{1,2}$ to align $s_1$ and $s_2$. 
This enables the checker to say that ``$s_2 \rightsquigarrow s_1$''.
Because $x_i$ is generated from $s_1$, the checker already
knows ``$s_1 \rightsquigarrow x_i$''. With that, the
checker now can say that ``$s_2 \rightsquigarrow s_1 \rightsquigarrow x_i$''.
From the checker point of view, it has verified that
$x_i$ can be transformed from {\it both} $s_1$ and $s_2$. 
Hence, $x_i$  becomes an {\it indifferentiable} paraphrase. 
Note that this indifferentiability is strictly from the
checker's viewpoint. In Figure~\ref{nonisomorphism}, 
after the random coin tossing if $x_i$ is chosen as $p$, 
then the queries are basically asking the oracle if
it can differentiate where $x_i$ comes from ($s_1$ or $s_2$). 

\subsection{Key points in our alignment-based approximation}
\label{sec04.2}

Note that an alignment scheme implemented by the checker
is not necessarily the alignment scheme the oracle uses
(if it actually even uses one)
for deciding semantic equivalence. 
Hence, it is expected that on a given set of 
oracle-claimed semantically
equivalent pairs, the checker
can only verify a subset of them. 
Let this subset be called a {\it provable} set.
An interesting aspect to observe in the experiment 
is the ratio of this provable set to the 
oracle-claimed set. 

In the above discussion, two alignments are mentioned:
$\rho$ and $\phi$. The former is directional and the
later is not.
In our checker, we implement a method $M_{\rho}$ and
based on it, two methods $M_{yes}$ and $M_{no}$ are implemented. 

It is interesting to see that in both yes and no cases, 
the checker relies on finding an alignment. This is similar to
the graph isomorphism check where in both cases, the checker
relies on finding a permutation: in the yes case a particular
one as the proof and in the no case, a random one.  
In the semantic equivalence context, this randomness
comes from oracle's ability to randomly generate paraphrases. 

\begin{figure*}[thb]
	\centering
	\includesvg[inkscapelatex=false, width=\textwidth]{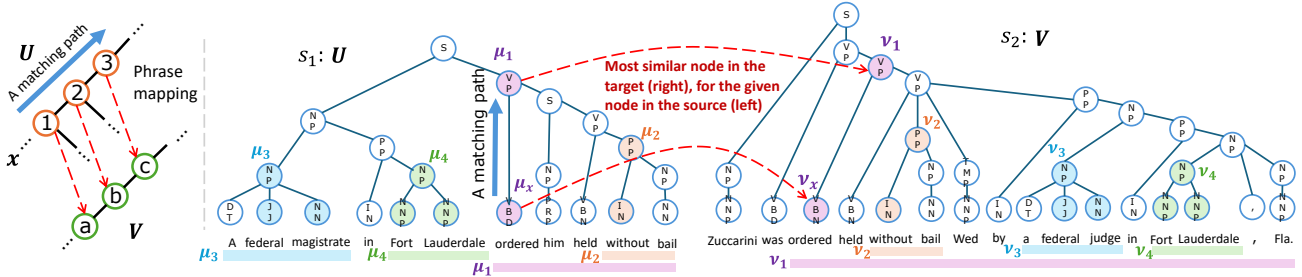}
	\vskip -0.1in
	\caption{Given two syntactic trees $U,V$ as source and target,
		respectively, a $\rho$-alignment is found between the two
		(in the example each phrase mapping, between a pair of nodes and its corresponding pair 
		of phrases, is colored the same). Searching for $\rho$-alignment is based on 
		matching a node $\mu$ in $U$ with its most similar node $\nu$ in $V$.
		The similarity from a node in $U$ to a node in $V$ is determined 
		by the BERT model \cite{devlin-etal-2019-bert}. 
		Each phrase mapping is a sub-tree containing
		a {\it matching path} of depth at least two from the sub-tree's root. 	\label{fig-2trees}}
\end{figure*}

\subsection{Finding $\rho$-alignments and $\phi$-alignments}
\label{sec04.3}

The method $M_{\rho}$ finds a $\rho$-alignment. 
We define $\rho$-alignment as the following. 
Let a sentence $s$ be represented as a sequence of tokens.
A {\it phrase} is a consecutive sub-sequence of tokens. 
Given two sentences $s_1, s_2$, a {\it phrase mapping} $\sigma$
is a mapping of two phrases $\sigma: p_1 \rightarrow p_2$ 
where $p_1 \in s_1$ and $p_2 \in s_2$. 
Phrase mappings are directional. 
We say there exists a $\rho$-alignment from $s_1$ 
to $s_2$ if there exists is a set of phrase mappings, 
$\{\sigma_1, \ldots, \sigma_k\}$ for some $k$,
such that all tokens in $s_1$, except for 
individual propositional words, are covered by a phrase mapping. 
For simplicity, we use $\rho_{s_1 \rightarrow s_2}$ to denote
a $\rho$-alignment from $s_1$ to $s_2$.  
%Note that phrase mapping and $\rho$-alignment are directional.  

\begin{algorithm}[htb]
	\caption{$M_{\rho}$: Find a $\rho$-alignment $\rho_{s_1 \rightarrow s_2}$}
	\label{alg:search}
	\begin{algorithmic}
		\STATE {\bfseries Input:} Two sentences $s_1$, $s_2$
		\STATE $U$, $V$ $\leftarrow$ $parse(s_1), parse(s_2)$, respectively
		\FOR{each node $\mu_i$ {\bfseries in} $U$}
			\STATE Find the most similar node $\nu_j$ in $V$; add edge $(\mu_i, \nu_j)$
		\ENDFOR
		\STATE $S_U \leftarrow \phi$
		\FOR{each leaf $x$ {\bfseries in} $U$}  
			\STATE $S_U \leftarrow S_U \cup matchingpath(U, V, x)$
		\ENDFOR
		\STATE Report success if $align(U, S_U)$ 
	\end{algorithmic}
\end{algorithm}

To determine if $s_1$ is $\rho$-aligned to $s_2$, 
we first obtain their constituency parsing trees $U, V$. 
This can be done with the core-NLP API \cite{qi2020stanza}. 
Then, we build a directional {\it matching graph} from nodes in
$U$ to nodes in $V$. For each node $\mu_i \in U$,
we find the most similar node $\nu_j$ in $V$ and we add
a directional edge $(\mu_i, \nu_j)$ to the matching graph.
The similarity score between two nodes are determined based
on their embeddings provided by a language model like BERT \cite{devlin-etal-2019-bert}. 

Based on the matching graph, Figure~\ref{fig-2trees} illustrates
how a phrase mapping is found and based on a set of
phrase mappings, how a $\rho$-alignment 
$\rho_{s_1 \rightarrow s_2}$ can be established. 
The process is described in Algorithm~\ref{alg:search} which
includes two main functions, $matchingpath()$ and 
$align()$. 

Figure~\ref{fig-2trees}'s left plot depicts
how $matchingpath(U,V,x)$ works. Starting from a leaf
$x \in U$, each node on the path is included with the
following check:
If $(\mu_1, \nu_1)$ is an edge in the matching graph, 
then $parent(\mu_1)$ 
is only included if (1) $(parent(\mu_1), parent(\nu_1))$
is also an edge and (2) 
$\mu_1$ has at least one sibling.
Further, a path must contain at least one additional
node other than $x$ and cannot include the root of $U$.
This is to ensure that the alignment is compositional and also avoid
matching the entire sentence, i.e. the
$\rho$-alignment represents a decomposition of the
original sentence. 
 
Suppose a matching path contains a sequence of nodes
$(x, \mu_1, \ldots, \mu_i)$. Node $\mu_i$ is the root 
of a sub-tree which covers at least two tokens. 
The sequence of tokens covered by the sub-tree is the
{\it phrase} corresponding to $\mu_i$.
Hence, the sub-tree roots of all matching paths represents
a collection of {\it matching phrases} in $s_1$.  
Note that the $V$'s nodes being matched by those roots
also form a collection of matching phrases in $s_2$.
This is depicted in the right plot of Figure~\ref{fig-2trees}.

In Algorithm~\ref{alg:search}, the set $S_U$ can 
therefore be seen as the collection of matching phrases. 
Thus, $align(U,S_U)$ basically is checking if all leaves
in $U$ (except for propositional words) are covered by a set of
matching phrases. 
In Figure~\ref{fig-2trees}, the set of mappings
$\rho_{s_1 \rightarrow s_2}=(\mu_3 \rightarrow \nu_3,
\mu_4 \rightarrow \nu_4, \mu_1 \rightarrow \nu_1)$ 
is a {\it potential} $\rho$-alignment from $s_1$ to $s_2$.

To verify that the alignment is indeed a $\rho$-alignment,
the $align()$ asks the oracle if each of the matching phrases
are semantically equivalent. 
In Figure~\ref{overview}, $\{q_1, \ldots, q_n\}$ 
correspond to asking the oracle those questions. 

Note that to find a $\rho_{s_2 \rightarrow s_1}$, we can
swap the two inputs in Algorithm~\ref{alg:search}.
Furthermore, in
the algorithm if we focus the coverage on $s_2$ rather
than $s_1$, we may find a reverse alignment.
Denote this as $\rho^*$ to indicate the coverage check
is on the second input sentence. 
Using the algorithm, we can then find four
types of $\rho$-alignment: $\rho_{s_1 \rightarrow s_2}$,
$\rho*_{s_1 \rightarrow s_2}$,
$\rho_{s_2 \rightarrow s_1}$, 
and
$\rho*_{s_2 \rightarrow s_1}$. 
Then, we say $\phi_{or}:s_1 \sim s_2$ if any one
of the four $\rho$-alignments can be established. 
We say $\phi_{and}:s_1 \sim s_2$ if among the four,
for both directions at least one $\rho$-alignment 
exists in that direction. 

%rtAB, rtBA, rpBA, rpAB
%so AND = (rtAB+rtBA)(rpBA+rpAB)
% = rtAB*rpBA + rtBA*rpBA + rtAB*apAB + rtBA*rpAB
% = remove t * remove p + t*p + t*p + t*p 

\section{Experiments}
\label{sec05}

For entity extraction, the findings are based on 5000 sentences
randomly sampled from the DOCRED dataset \cite{yao-etal-2019-docred}. 
In addition, we use 500 sentences randomly
sampled from the RISC-V unprivileged ISA specification 
\cite{riscv-unpriviledged} for comparison purpose. 

For paraphrasing, the experiments are based on the MSR Paraphrase 
corpus \cite{dolan-brockett-2005-automatically}. 
In all experiments, we treat GPT3.5 \cite{GPT3.5} as our oracle.
Detail regarding the experiment setup and the 
prompts to GPT3.5 can be found in the Appendix
sections~\ref{app01} and~\ref{app02}. 

\subsection{Results on entity extraction} 
\label{sec05.01}

As mentioned in section~\ref{sec03.2}, the linearity test
relies on the synonym generator $syn$ and the
entity replacement $R$. For $syn$, we rely on GPT3.5
for providing a synonym for a given entity. 
For $R$, we simply keep the original sentence by
replacing entities with their synonyms according to
the characteristic vector. 
%(see Appendix section~\ref{app01} for detail). 

For each sentence, we requests GPT3.5 to perform
entity extraction. In our experiments 
this request is repeated 11 times
and the most frequent set of entities
is used as the entity set for running the subsequent linearity test. 
For each sentence, five tests are run. 
Each test requires random sampling of two
characteristic vectors $x,y$, resulting in
three $R$-transformed sentences
(for ``$x$'', ``$y$'' and ``$x+y$''). 
The entity extraction on each $R$-transformed sentence
is repeated 11 times. 
Therefore, in total the acceptance/rejection decision on
each sentence is based on 165 calls to GPT3.5.

\begin{figure}[htb]
	\centering
	\includesvg[inkscapelatex=false, width=3.2in]{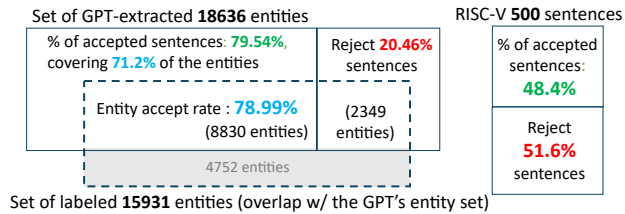}
		\vskip -0.1in
	\caption{Linearity test results on 5000 DOCRED sentences.
	Two entity sets are considered: GPT-extracted set with 18636
    entities and human-labeled set with 15931 entities. 
    The two sets overlap on about 70\% ($=\frac{8830+2349}{15931}$) 
    of the labeled set. On the RISC-V 500 sentences, the 
    acceptance rate is much lower (48.4\% vs. 79.54\%). }
	\label{linearity}
%	\vskip -0.1in
\end{figure}

Because entity extraction on the three $R$-transformed
sentences each has 11 results, one linearity test 
can consider $11^3$ combinations. 
In our experiments we consider any of the combinations
passing the test as passing. However, if we focus on 
those accepted sentence and calculate a passing rate 
across all combinations, then on average we observe 
the passing rate is about 38.48\%.

Figure~\ref{linearity} summarizes the results. 
There are several observations: 
(1)   79.54\% of the 5000 sentences accepted 
by the checker which cover about
71.2\% of the total 18636 entities GPT3.5 extracted. 
(2) Among the 15931 entities labeled in the dataset,
about 70\% coincide with the GPT extracted set. 
(3) There is a slightly higher acceptance rate,
78.99\%, on entities if they are in both the GPT set
and the labeled set. 

From the experiment, we can conclude that on at least 
20\% of the sentences, the GPT3.5's entity extraction
is not linear. Examples on what might cause this 
non-linearity extraction are discussed further in
the Appendix (section~\ref{app.entity.fails}).
 
For the 500 RISC-V sentences, the per-sentence acceptance
rate is 48.4\%, much lower than the 79.54\% mentioned above. 
The per-entity acceptance rate is 34\%, comparing to
71.2\% above. This indicates that entity extraction by
GPT3.5 on the RISC-V specification is much more ``complex''
than on the DOCRED dataset. Examples on this complexity
increase is discussed in detail in Appendix
section~\ref{app.entity.riscv}. 

Table~\ref{correlations} shows Spearman correlation between
different quantities measured in the experiments. 
Observe that for sentences from
the DOCRED dataset, the passing rate
of a linearity test ($A_{rate}$) is highly correlated 
(0.859) to
the minimum consistency ($Con_{rs}$) observed in the repeated
calls to GPT3.5 for executing the test. 
Similarly, we observe a high correlation
between $A_{rate}$ and the consistency measure $Con_o$.
In contrast, $A_{rate}$ is not as correlated 
($-$0.414) to the number
of entities being tested ($\# E$). 

The results on the RISC-V sentences
further shows how GPT3.5 behaves differently on them. 
For example, the correlation between $A_{rate}$ and $Con_{rs}$
is lower (but remains high), and yet its correlation to $\# E$ 
becomes stronger ($-0.651$). 

\begin{table}[t]
	\setlength{\tabcolsep}{4pt}
	\caption{Correlations: 
	$Con_o$: out of 11 requests on the original sentence,
	the most frequent answer is denoted as $E$. $Con_o$ is
	the number of times $E$ is the answer.  
	$Con_{rs}$: there are 15 $R$-transformed sentences. 
	On each sentence, we can calculate a $Con_r$ similar to
	$Con_o$ above. $Con_{rs}$ is the smallest number 
	among them. 
	$A_{test}$: the number of passing tests for the five tests on a given sentence.
	$A_{rate}$: the passing rate across all combinations
	considered by the five tests. 
	$\# E$: denote the number of entities. 
	The ``$a$ / $b$'' corresponds to DOCRED/RISC-V, respectively.
    \label{correlations} }
	\vskip -0.1in
	\begin{center}
		\begin{small}
				\begin{tabular}{cccc}
					\toprule
					 $a$ / $b$ & $Con_o$ & $Con_{rs}$ & $\# E$ \\
					\midrule
				 	$A_{test}$    & 0.447/0.511 & 0.422/0.539 & $-0.399$/$-0.61$ \\
					$A_{rate}$    & 0.789/0.648 & {\bf 0.859}/{\bf 0.721} & $-${\bf 0.414}/$-${\bf 0.651} \\
					$\# E$       & $-0.264$/$-0.426$ & $-0.302$/$-0.453$ & 1.0/1.0 \\
					\bottomrule
				\end{tabular}
		\end{small}
	\end{center}
	\vskip -0.1in
\end{table}

\subsection{Results on paraphrase decision} 
\label{sec05.02}

In this work, we use the term {\it semantic equivalence} in
the queries to GPT3.5. 
In practice, 
``semantic equivalence'' can really means ``semantic near-equivalence''
that a pair of sentences ideally entail each other but often
might have some minor mismatches in their content \cite{dolan-brockett-2005-automatically}.
Deciding whether or not the difference in content  
is significant enough to make the two not semantically equivalent, can
become a personal judgment call. 
In the MSR Paraphrase corpus, the label annotators made 
these judgment calls. When we ask an oracle with paraphrase decisions,
the oracle make its own judgment calls. 

\subsubsection{Sanity checks}
\label{sec05.02.0}

The MSRP dataset contains pairs of sentences, say
$(s,pr)$, to be considered for {\it potential} semantic 
equivalence. 
As a first sanity check, we randomly sampled 500 
pairs and repeatedly asked GPT3.5 five times whether or not 
each was semantically equivalent. Among them, the GPT3.5's
answers were inconsistent only on 59 pairs. Hence, if
we relied on self-consistency for deciding 
acceptance/rejection of GPT3.5's answers, we would 
reject only 11.8\% of them. 

As a second sanity check, we randomly sampled 500 sentences
and asked GPT3.5 to generate a paraphrase
for each. Then, we repeated the same experiment
above. In this case, the GPT3.5 was inconsistent only on
four pairs, or 0.8\%. This shows that inconsistency is
not an effective way to check on GPT3.5 own generated
paraphrases.   
As a third sanity check, we construct a dataset of 500 semantically
non-equivalent pairs, each by randomly sampling two sentences 
only from the $s$ part in the MSRP dataset. 
For all 500 pairs, GPT3.5 says no on their
semantic equivalence.

Then, we ran our checker to prove semantic equivalence on 
all of them
where the equivalence was based on the more relaxed $\phi_{or}$ alignment.
The checker could {\it not} generate a proof on any. 
Specifically, in the $align()$ function in Algorithm~\ref{alg:search},
only 61 potentially aligned pairs had queries sent to GPT3.5 and failed
afterward. This means that for the rest of 439 pairs the
prover failed without even asking GPT3.5. 
In addition, we ran a test similar to what depicted in
Figure~\ref{nonisomorphism} where $p$ is randomly
chosen from 10 paraphrases (5 from $s_1$, 5 from $s_2$) instead of
finding an indifferentiable~$p$.
The test accepts all the 500 GPT3.5's no answers. 

\subsubsection{Results from the proof perspective}
\label{sec05.02.1}

The sanity check confirms that for those obvious non-equivalent
pairs, GPT3.5, our equivalence prover, and a random
test can all agree with each other. However, as we will see that
this is not the case for those pairs from the MSRP dataset. 
Note that for many MSRP pairs, even though
they are labeled as no, this ``no'' can be much closer to
being a ``yes'' than those pairs used in our sanity check. 

\begin{table}[ht]
	\vskip -0.1in
	\setlength{\tabcolsep}{4pt}
	\caption{Yes/No results given by GPT3.5 on 5000 sentence
		pairs from MSRP. In total, GPT3.5 agrees on 74.84\% of the labels.
	\label{MSRPvsGPT}}
	\begin{center}
		\begin{small}
			\begin{tabular}{cccc}
				\toprule
				\multicolumn{2}{c}{3454 Labeled Yes} & \multicolumn{2}{c}{1546 Labeled No} \\
				\midrule
				GPT3.5 Yes & GPT3.5 No & GPT3.5 Yes & GPT3.5 No \\
				77.65\% & 22.35\% & 31.44\% & 68.56\% \\
				\bottomrule
			\end{tabular}
		\end{small}
	\end{center}
	\vskip -0.1in
\end{table}

\begin{table}[ht]
	\setlength{\tabcolsep}{4pt}
	\caption{\% of pairs accepted by our Yes-checker. 
		The checker's prover agrees with GPT3.5 slightly more,
		i.e. the \% on GPT3.5 Yes cases is slightly higher 
		than the \% on Labeled Yes cases. The \% shown under
		those No cases are for reference only, as in practice our
		Yes-checker would never run on them. \label{MSRPvsGPTvsChecker}}
	\vskip -0.1in
	\begin{center}
		\begin{small}
			\begin{tabular}{cccc}
				\toprule
				\multicolumn{2}{c}{Based on $\phi_{or}$} 
				& \multicolumn{2}{c}{Based on $\phi_{and}$} \\
				\midrule
				Labeled Yes & Label No & Labeled Yes & Label No \\
				
				53.97\% & 25.86\% & 31.77\% & 9.66\% \\
				\midrule \midrule
				GPT3.5 Yes & GPT3.5 No & GPT3.5 Yes & GPT3.5 No \\
				
				57.02\% & 24.45\% & 32.7\% & 11.20\% \\
				\bottomrule
			\end{tabular}
		\end{small}
	\end{center}
	\vskip -0.1in
\end{table}

Table~\ref{MSRPvsGPT} compares the MSRP labels
with the yes/no decisions given by GPT3.5. As seen, GPT3.5 only
agrees on 74.84\% of the labels 
(for detail see section~\ref{app02.2}).
Then, Table~\ref{MSRPvsGPTvsChecker} shows the acceptance rates
by our Yes-checker on those pairs. 
It is important to note that under our oracle-checker scheme,
the checker {\it would not} even try to prove equivalence on those
``GPT3.5's No'' cases. The no answers are checked separately with
our non-equivalence test. Nevertheless, we include the rates
on those ``GPT3.5's No'' cases to explain the observations below
(for detail see section~\ref{app02.3}).

We can make several observations on Table~\ref{MSRPvsGPTvsChecker}:
(1) For both $\phi_{or}$ and $\phi_{and}$, the rate on
``Yes'' is much higher than the rate on ``No''. 
These results confirm the validity of our prover. 
(2) $\phi_{and}$ involves a more stringent condition
to establish equivalence and hence, its ``Yes'' acceptance rates 
are lower than $\phi_{or}$'s. 
(3) If we focus on the difference between the rate for ``Yes'' 
and the rate for ``No'' in each respective case, 
and consider a larger difference as more effective, then the
effectiveness is comparable between checking GPT3.5's decisions 
and checking label annotators' decisions. 
(4) 
Even though the previous result
shows that GPT3.5 disagrees on 25.16\% of the annotators' labels, 
for those ``Yes'' cases 
we see the results are comparable between ``GPT3.5 Yes'' and
``Labeled Yes'', i.e. 
from our checker's perspective, the acceptance rates
of their ``Yes'' decisions are similar. 
(5) Finally, our checker can accept a reasonably large
percentages of ``Yes'' cases. This is useful in practice 
if we desire to select semantically equivalent pairs
of sentences when we do not completely trust 
GPT3.5-provided decisions. 
 
\begin{table}[ht]
	\setlength{\tabcolsep}{3pt}
	\caption{Checker's SE-provable \% on 500 GPT3.5-paraphrased
		sentence pairs. GPT3.5 disagrees on 5 of its own paraphrases. \label{CheckervsGPT}}
	\vskip -0.1in
	\begin{center}
		\begin{small}
			\begin{tabular}{cccc}
				\toprule
				\multicolumn{2}{c}{Yes:495} & \multicolumn{2}{c}{No:5} \\
				\midrule
				Based on $\phi_{or}$ & Based on $\phi_{and}$ &
				Based on $\phi_{or}$ & Based on $\phi_{and}$ \\
				62.5\% & 45.34\% & 60\% &  60\%\\
				\bottomrule
			\end{tabular}
		\end{small}
	\end{center}
	\vskip -0.1in
\end{table}

Table~\ref{CheckervsGPT} shows results on 500 pairs of GPT3.5-generated paraphrases.
GPT3.5 answers ``Yes'' on the semantic equivalence for 495 of them.
As shown, our checker's success rates on these ``Yes'' decisions are much higher than
those in Table~\ref{MSRPvsGPTvsChecker} before. In particular,
the rate of $\phi_{and}$ (45.34\%) is much higher than before. 
This indicates that GPT3.5-generated paraphrases have a higher chance
satisfying the stronger alignment property, $\phi_{and}$.  
The rates on the ``No'' cases are not very meaningful as there are
only five of them. Appendix section~\ref{fiveNOcases} provides
more detail about them.

\subsubsection{Results from the trust perspective}

\begin{table}[t]
	\setlength{\tabcolsep}{3pt}
	\caption{On 1832 ``GPT3.5-No'' cases, show the \# of cases
		in each of the four categories  
		($AI$, $\bar{A}I$, $A\bar{I}$, $\bar{A}\bar{I}$) 
		and their \%  
		accepted by the checker in {\it that} category (shown as the
		\% below the \#).
	Refer to Figure~\ref{nonisomorphism}.
	The first letter ``$A$'' indicates a found
	$\phi$ such that $\phi:s_1 \sim s_2$. 
	``$\bar{A}$'' means otherwise. 
	The second letter ``$I$'' indicates a found indifferentiable $p$
	and ``$\bar{I}$'' indicates otherwise. 	\label{Trust}
	}
	\vskip -0.1in
	\begin{center}
		\begin{small}
			\begin{tabular}{cccc|cccc}
				\toprule
				\multicolumn{4}{c}{Based on $\phi_{or}$} 
				& \multicolumn{4}{c}{Based on $\phi_{and}$} \\
				\midrule
				$AI$ & $\bar{A}I$ & $A\bar{I}$ & $\bar{A}\bar{I}$ &
				$AI$ & $\bar{A}I$ & $A\bar{I}$ & $\bar{A}\bar{I}$
				\\
				383 & 475 & 32 & 942 &
				152 & 233 & 38 & 1409 \\
				\midrule
				29.2\% & 45.9\% & 62.5\% & 81.9\% & 
				21.7\% & 34.8\% & 23.7\% & 71\% \\
				\bottomrule
			\end{tabular}
		\end{small}
	\end{center}
	\vskip -0.1in
\end{table}

It should be noted that for many of the MSRP pairs, 
deciding their semantic equivalence can be rather subjective.
Hence, there is no definite ``correct'' answer on those cases.
%(see discussion in the Appendix, sections~\ref{app02.2} and~\ref{app02.3}). 
The effect of such cases on our ``Yes''
checker is limited because the checker only accepts a yes
answer when it finds its own proof of
semantic equivalence. For the ``No'' checker
the effect is less controllable because of the randomness
involved in the test. Nevertheless, it is important to note
that when the checker rejects a no answer, it does have the
following ``triangular'' evidence: GPT3.5 says that 
a paraphrase $p$ (whenever possible, an indifferentiable 
paraphrase) is semantically equivalent to both $s_1$ and $s_2$.
This contradicts to the the original claim that 
$s_1$ and $s_2$ are not semantically equivalent
(see discussion in the Appendix, sections~\ref{app02.4}).

Table~\ref{Trust} shows the results on the ``No'' cases. 
We divide the results into four categories using two
letters, ``$A$'' (alignment) and ``$I$'' (indifferentiable)
to indicate them. The experiment was based on
selecting one test (a $p$ in Figure~\ref{nonisomorphism})
and accepted the no answer if the result passed the test
and otherwise, rejected it. The table shows the acceptance rates. 
The four categories were run in priority order from 
left to right, i.e. if a case is found in the $AI$
category, it will be tested under $AI$ first. A case is tested
under $\bar{A}\bar{I}$ only when it does not belong to
all the previous three categories. 

The $A\bar{I}$ results are less meaningful as they
have much less number of cases. Comparing $\bar{A}\bar{I}$
(in $\bar{I}$, a random paraphrase $r$ is selected)
to the first two combined (in both $AI$ and $\bar{A}{I}$,
an indifferentiable paraphrase $p$ is used), we see that
using an indifferentiable $p$ has a much higher rejection rate than $r$. 
This indicates the effectiveness of involving 
$\phi$-indifferentiability to find evidence for exposing
contradiction in the ``No'' claims.  
In that regard, $\phi_{or}$ is less effective than
$\phi_{and}$, as expected. However, $\phi_{and}$ is applicable
to less number of cases. This shows a tradeoff between
the two.

\section{Related work} 
\label{sec06}

There are many works on characterization and trustworthiness of LLMs. 
They can be viewed in three contexts:
%Following a general benchmark-driven strategy, 
%the basis for preferred behavior is captured by a dataset 
%which is used to evaluate the performance of LLMs. 
(1) Fine tuning and evaluating on a collection of datasets is a popular approach 
for improving LLMs performance \cite{wei2022finetuned} \cite{ouyang2022training} 
\cite{openai2023gpt4}, \cite{touvron2023llama}. 
Example benchmarks include those to evaluate human-GPT differences,
broadly \cite{guo2023close} and specifically in education \cite{herbold2023ai}, 
\cite{PhysRevPhysEducRes.19.010132} and those 
on overall hallucination tendencies \cite{culture10319443}, \cite{rawte2023troubling} \cite{Johnson2023-vi}. 
(2) There are works by curating a dataset specifically 
for the purpose of improving the models 
in the context of enabling preference or ranking to the output results 
or flagging potentially harmful results. 
Often the product is an auxiliary model, for example, 
as mathematical reward models \cite{lightman2023lets}, 
for improvement on code vulnerability \cite{bhatt2023purple}, and 
for enhancement on illegal topic detection \cite{inan2023llama}.
(3) Self-consistency can be an important method for evaluating a LLM. 
A recent work uses self-consistency \cite{wang2023selfconsistency}
to improve performance,
where a followup expands into a universal consistency check 
\cite{chen2023universal}. Auxiliary models can also be reintroduced to 
self-consistency as alternatives to voting \cite{manakul2023selfcheckgpt}.
In addition to the above works, there is also work focusing on
the effect of the prompts by evaluating if the prompts are 
at fault for failure cases \cite{lin2022truthfulqa}.

Our work takes a direction different from those reviewed above,
and provides a new framework for evaluating trustworthiness of LLMs. 
Our framework can be complementary to a benchmark-driven approach, 
and the proposed oracle-checker scheme is inspired by 
the theoretical works established decades ago. 
Our work can be useful when one needs to enforce
a subjective view on trustworthiness or when one works on  
a domain where the data characteristics are drastically 
different from public datasets.
 
%Our approach is based on the 
%and the focus is on realizing their ideas in practice by treating a 
%LLM as an oracle. 

\section{Conclusion}
\label{sec07}

Inspired by the foundation works \cite{BlumKanna95}\cite{BlumLubyRubinfeld90},
this work presents an oracle-checker scheme with three checking strategies:
{\it property}, {\it proof}, and {\it trust}. For each strategy, we show
how a practical checker can be implemented by viewing GPT3.5 as the oracle. 
We highlight key aspects that need attention when designing a checker.
The theoretical field includes a rich set of results 
\cite{Dana10}\cite{Goldreich17}
and this work only relies on some of the earliest ideas.

Our oracle-checker scheme can become handy when articulating
a subjective view is difficult in advance or when 
one tries to assess the trustworthiness of a LLM based on the view. 
Throughout our experiments, we observed interesting findings
which were not entirely anticipated, thanks to the checkers.
We do not claim that our current checkers are optimal. 
Their optimization are left to future work. 

\newpage

\section{Impact Statements}

This paper presents work whose goal is to improve on the trustworthiness 
of using a LLM. 
The work represents a first step toward a new direction
and the focus is limited to studying scientific aspects. 
While there might be potential societal consequences 
of the work in the future, none which we are aware of
can be specifically highlighted here.

\bibliography{custom,thomaslitreview,anthology}

\begin{thebibliography}{31}
\providecommand{\natexlab}[1]{#1}
\providecommand{\url}[1]{\texttt{#1}}
\expandafter\ifx\csname urlstyle\endcsname\relax
  \providecommand{\doi}[1]{doi: #1}\else
  \providecommand{\doi}{doi: \begingroup \urlstyle{rm}\Url}\fi

\bibitem[Arase \& Tsujii(2020)Arase and
  Tsujii]{arase-tsujii-2020-compositional}
Arase, Y. and Tsujii, J.
\newblock Compositional phrase alignment and beyond.
\newblock In Webber, B., Cohn, T., He, Y., and Liu, Y. (eds.),
  \emph{Proceedings of the 2020 Conference on Empirical Methods in Natural
  Language Processing (EMNLP)}, pp.\  1611--1623, Online, November 2020.
  Association for Computational Linguistics.
\newblock \doi{10.18653/v1/2020.emnlp-main.125}.
\newblock URL \url{https://aclanthology.org/2020.emnlp-main.125}.

\bibitem[Bhatt et~al.(2023)Bhatt, Chennabasappa, Nikolaidis, Wan, Evtimov,
  Gabi, Song, Ahmad, Aschermann, Fontana, Frolov, Giri, Kapil, Kozyrakis,
  LeBlanc, Milazzo, Straumann, Synnaeve, Vontimitta, Whitman, and
  Saxe]{bhatt2023purple}
Bhatt, M., Chennabasappa, S., Nikolaidis, C., Wan, S., Evtimov, I., Gabi, D.,
  Song, D., Ahmad, F., Aschermann, C., Fontana, L., Frolov, S., Giri, R.~P.,
  Kapil, D., Kozyrakis, Y., LeBlanc, D., Milazzo, J., Straumann, A., Synnaeve,
  G., Vontimitta, V., Whitman, S., and Saxe, J.
\newblock Purple llama cyberseceval: A secure coding benchmark for language
  models, 2023.

\bibitem[Blum \& Kannan(1995)Blum and Kannan]{BlumKanna95}
Blum, M. and Kannan, S.
\newblock Designing programs that check their work.
\newblock \emph{Journal of ACM}, 42\penalty0 (1):\penalty0 269--291, 1995.

\bibitem[Blum et~al.(1990)Blum, Luby, and Rubinfeld]{BlumLubyRubinfeld90}
Blum, M., Luby, M., and Rubinfeld, R.
\newblock Self-testing/correcting with applications to numerical problems.
\newblock In \emph{Proceedings of the 22nd Annual ACM Symposium on Theory of
  Computing (STOC)}, pp.\  73--83, 1990.

\bibitem[Chen et~al.(2023)Chen, Aksitov, Alon, Ren, Xiao, Yin, Prakash, Sutton,
  Wang, and Zhou]{chen2023universal}
Chen, X., Aksitov, R., Alon, U., Ren, J., Xiao, K., Yin, P., Prakash, S.,
  Sutton, C., Wang, X., and Zhou, D.
\newblock Universal self-consistency for large language model generation, 2023.

\bibitem[Devlin et~al.(2019)Devlin, Chang, Lee, and
  Toutanova]{devlin-etal-2019-bert}
Devlin, J., Chang, M.-W., Lee, K., and Toutanova, K.
\newblock {BERT}: Pre-training of deep bidirectional transformers for language
  understanding.
\newblock In \emph{Proceedings of the 2019 Conference of the North {A}merican
  Chapter of the Association for Computational Linguistics: Human Language
  Technologies, Volume 1 (Long and Short Papers)}, pp.\  4171--4186,
  Minneapolis, Minnesota, June 2019. Association for Computational Linguistics.
\newblock \doi{10.18653/v1/N19-1423}.
\newblock URL \url{https://aclanthology.org/N19-1423}.

\bibitem[Dolan \& Brockett(2005)Dolan and
  Brockett]{dolan-brockett-2005-automatically}
Dolan, W.~B. and Brockett, C.
\newblock Automatically constructing a corpus of sentential paraphrases.
\newblock In \emph{Proceedings of the Third International Workshop on
  Paraphrasing ({IWP}2005)}, 2005.
\newblock URL \url{https://aclanthology.org/I05-5002}.

\bibitem[Goldreich(2017)]{Goldreich17}
Goldreich, O. (ed.).
\newblock \emph{Introduction to Property Testing}.
\newblock Cambridge University Press, 2017.

\bibitem[Guo et~al.(2023)Guo, Zhang, Wang, Jiang, Nie, Ding, Yue, and
  Wu]{guo2023close}
Guo, B., Zhang, X., Wang, Z., Jiang, M., Nie, J., Ding, Y., Yue, J., and Wu, Y.
\newblock How close is chatgpt to human experts? comparison corpus, evaluation,
  and detection, 2023.

\bibitem[Herbold et~al.(2023)Herbold, Hautli-Janisz, Heuer, Kikteva, and
  Trautsch]{herbold2023ai}
Herbold, S., Hautli-Janisz, A., Heuer, U., Kikteva, Z., and Trautsch, A.
\newblock Ai, write an essay for me: A large-scale comparison of human-written
  versus chatgpt-generated essays, 2023.

\bibitem[Inan et~al.(2023)Inan, Upasani, Chi, Rungta, Iyer, Mao, Tontchev, Hu,
  Fuller, Testuggine, and Khabsa]{inan2023llama}
Inan, H., Upasani, K., Chi, J., Rungta, R., Iyer, K., Mao, Y., Tontchev, M.,
  Hu, Q., Fuller, B., Testuggine, D., and Khabsa, M.
\newblock Llama guard: Llm-based input-output safeguard for human-ai
  conversations, 2023.

\bibitem[Johnson et~al.(2023)Johnson, Goodman, Patrinely, Stone, Zimmerman,
  Donald, Chang, Berkowitz, Finn, Jahangir, Scoville, Reese, Friedman,
  Bastarache, van~der Heijden, Wright, Carter, Alexander, Choe, Chastain, Zic,
  Horst, Turker, Agarwal, Osmundson, Idrees, Kieman, Padmanabhan, Bailey,
  Schlegel, Chambless, Gibson, Osterman, and Wheless]{Johnson2023-vi}
Johnson, D., Goodman, R., Patrinely, J., Stone, C., Zimmerman, E., Donald, R.,
  Chang, S., Berkowitz, S., Finn, A., Jahangir, E., Scoville, E., Reese, T.,
  Friedman, D., Bastarache, J., van~der Heijden, Y., Wright, J., Carter, N.,
  Alexander, M., Choe, J., Chastain, C., Zic, J., Horst, S., Turker, I.,
  Agarwal, R., Osmundson, E., Idrees, K., Kieman, C., Padmanabhan, C., Bailey,
  C., Schlegel, C., Chambless, L., Gibson, M., Osterman, T., and Wheless, L.
\newblock Assessing the accuracy and reliability of {AI-generated} medical
  responses: An evaluation of the {chat-GPT} model.
\newblock \emph{Res. Sq.}, February 2023.

\bibitem[Jurafsky \& Martin(2023)Jurafsky and Martin]{stanford-etextbook}
Jurafsky, D. and Martin, J.~H.
\newblock \emph{Speech and Language Processing: An Introduction to Natural
  Language Processing, Computational Linguistics, and Speech Recognition}.
\newblock 2023.
\newblock URL \url{https://web.stanford.edu/~jurafsky/slp3/}.

\bibitem[Kortemeyer(2023)]{PhysRevPhysEducRes.19.010132}
Kortemeyer, G.
\newblock Could an artificial-intelligence agent pass an introductory physics
  course?
\newblock \emph{Phys. Rev. Phys. Educ. Res.}, 19:\penalty0 010132, May 2023.
\newblock \doi{10.1103/PhysRevPhysEducRes.19.010132}.
\newblock URL
  \url{https://link.aps.org/doi/10.1103/PhysRevPhysEducRes.19.010132}.

\bibitem[Krallinger et~al.(2015)Krallinger, Rabal, Leitner, Vazquez, Salgado,
  Lu, Leaman, Lu, Ji, Lowe, Sayle, Batista-Navarro, Rak, Huber,
  Rockt{\"a}schel, Matos, Campos, Tang, Xu, Munkhdalai, Ryu, Ramanan, Nathan,
  {\v{Z}}itnik, Bajec, Weber, Irmer, Akhondi, Kors, Xu, An, Sikdar, Ekbal,
  Yoshioka, Dieb, Choi, Verspoor, Khabsa, Giles, Liu, Ravikumar, Lamurias,
  Couto, Dai, Tsai, Ata, Can, Usi{\'e}, Alves, Segura-Bedmar, Mart{\'i}nez,
  Oyarzabal, and Valencia]{Krallinger2015}
Krallinger, M., Rabal, O., Leitner, F., Vazquez, M., Salgado, D., Lu, Z.,
  Leaman, R., Lu, Y., Ji, D., Lowe, D.~M., Sayle, R.~A., Batista-Navarro,
  R.~T., Rak, R., Huber, T., Rockt{\"a}schel, T., Matos, S., Campos, D., Tang,
  B., Xu, H., Munkhdalai, T., Ryu, K.~H., Ramanan, S.~V., Nathan, S.,
  {\v{Z}}itnik, S., Bajec, M., Weber, L., Irmer, M., Akhondi, S.~A., Kors,
  J.~A., Xu, S., An, X., Sikdar, U.~K., Ekbal, A., Yoshioka, M., Dieb, T.~M.,
  Choi, M., Verspoor, K., Khabsa, M., Giles, C.~L., Liu, H., Ravikumar, K.~E.,
  Lamurias, A., Couto, F.~M., Dai, H.-J., Tsai, R. T.-H., Ata, C., Can, T.,
  Usi{\'e}, A., Alves, R., Segura-Bedmar, I., Mart{\'i}nez, P., Oyarzabal, J.,
  and Valencia, A.
\newblock The chemdner corpus of chemicals and drugs and its annotation
  principles.
\newblock \emph{Journal of Cheminformatics}, 7\penalty0 (1):\penalty0 S2, Jan
  2015.
\newblock ISSN 1758-2946.
\newblock \doi{10.1186/1758-2946-7-S1-S2}.
\newblock URL \url{https://doi.org/10.1186/1758-2946-7-S1-S2}.

\bibitem[Lightman et~al.(2023)Lightman, Kosaraju, Burda, Edwards, Baker, Lee,
  Leike, Schulman, Sutskever, and Cobbe]{lightman2023lets}
Lightman, H., Kosaraju, V., Burda, Y., Edwards, H., Baker, B., Lee, T., Leike,
  J., Schulman, J., Sutskever, I., and Cobbe, K.
\newblock Let's verify step by step, 2023.

\bibitem[Lin et~al.(2022)Lin, Hilton, and Evans]{lin2022truthfulqa}
Lin, S., Hilton, J., and Evans, O.
\newblock Truthfulqa: Measuring how models mimic human falsehoods, 2022.

\bibitem[Manakul et~al.(2023)Manakul, Liusie, and
  Gales]{manakul2023selfcheckgpt}
Manakul, P., Liusie, A., and Gales, M. J.~F.
\newblock Selfcheckgpt: Zero-resource black-box hallucination detection for
  generative large language models, 2023.

\bibitem[McIntosh et~al.(2023)McIntosh, Liu, Susnjak, Watters, Ng, and
  Halgamuge]{culture10319443}
McIntosh, T.~R., Liu, T., Susnjak, T., Watters, P., Ng, A., and Halgamuge,
  M.~N.
\newblock A culturally sensitive test to evaluate nuanced gpt hallucination.
\newblock \emph{IEEE Transactions on Artificial Intelligence}, pp.\  1--13,
  2023.
\newblock \doi{10.1109/TAI.2023.3332837}.

\bibitem[OpenAI()]{GPT3.5}
OpenAI.
\newblock Gpt-3.5-turbo.
\newblock URL \url{https://platform.openai.com/docs/models/gpt-3-5}.

\bibitem[OpenAI \& et. al(2023)OpenAI and et. al]{openai2023gpt4}
OpenAI and et. al.
\newblock Gpt-4 technical report, 2023.

\bibitem[Ouyang et~al.(2022)Ouyang, Wu, Jiang, Almeida, Wainwright, Mishkin,
  Zhang, Agarwal, Slama, Ray, Schulman, Hilton, Kelton, Miller, Simens, Askell,
  Welinder, Christiano, Leike, and Lowe]{ouyang2022training}
Ouyang, L., Wu, J., Jiang, X., Almeida, D., Wainwright, C.~L., Mishkin, P.,
  Zhang, C., Agarwal, S., Slama, K., Ray, A., Schulman, J., Hilton, J., Kelton,
  F., Miller, L., Simens, M., Askell, A., Welinder, P., Christiano, P., Leike,
  J., and Lowe, R.
\newblock Training language models to follow instructions with human feedback,
  2022.

\bibitem[Qi et~al.(2020)Qi, Zhang, Zhang, Bolton, and Manning]{qi2020stanza}
Qi, P., Zhang, Y., Zhang, Y., Bolton, J., and Manning, C.~D.
\newblock Stanza: A {Python} natural language processing toolkit for many human
  languages.
\newblock In \emph{Proc. 58th ACL: System Demonstrations}, 2020.
\newblock URL \url{https://nlp.stanford.edu/pubs/qi2020stanza.pdf}.

\bibitem[Rawte et~al.(2023)Rawte, Chakraborty, Pathak, Sarkar, Tonmoy, Chadha,
  Sheth, and Das]{rawte2023troubling}
Rawte, V., Chakraborty, S., Pathak, A., Sarkar, A., Tonmoy, S. M. T.~I.,
  Chadha, A., Sheth, A.~P., and Das, A.
\newblock The troubling emergence of hallucination in large language models --
  an extensive definition, quantification, and prescriptive remediations, 2023.

\bibitem[Ron(2010)]{Dana10}
Ron, D.
\newblock \emph{Algorithmic and Analysis Techniques in Property Testing}.
\newblock Now Foundations and Trends, 2010.

\bibitem[Touvron et~al.(2023)Touvron, Martin, Stone, Albert, Almahairi, Babaei,
  Bashlykov, Batra, Bhargava, Bhosale, Bikel, Blecher, Ferrer, Chen, Cucurull,
  Esiobu, Fernandes, Fu, Fu, Fuller, Gao, Goswami, Goyal, Hartshorn, Hosseini,
  Hou, Inan, Kardas, Kerkez, Khabsa, Kloumann, Korenev, Koura, Lachaux, Lavril,
  Lee, Liskovich, Lu, Mao, Martinet, Mihaylov, Mishra, Molybog, Nie, Poulton,
  Reizenstein, Rungta, Saladi, Schelten, Silva, Smith, Subramanian, Tan, Tang,
  Taylor, Williams, Kuan, Xu, Yan, Zarov, Zhang, Fan, Kambadur, Narang,
  Rodriguez, Stojnic, Edunov, and Scialom]{touvron2023llama}
Touvron, H., Martin, L., Stone, K., Albert, P., Almahairi, A., Babaei, Y.,
  Bashlykov, N., Batra, S., Bhargava, P., Bhosale, S., Bikel, D., Blecher, L.,
  Ferrer, C.~C., Chen, M., Cucurull, G., Esiobu, D., Fernandes, J., Fu, J., Fu,
  W., Fuller, B., Gao, C., Goswami, V., Goyal, N., Hartshorn, A., Hosseini, S.,
  Hou, R., Inan, H., Kardas, M., Kerkez, V., Khabsa, M., Kloumann, I., Korenev,
  A., Koura, P.~S., Lachaux, M.-A., Lavril, T., Lee, J., Liskovich, D., Lu, Y.,
  Mao, Y., Martinet, X., Mihaylov, T., Mishra, P., Molybog, I., Nie, Y.,
  Poulton, A., Reizenstein, J., Rungta, R., Saladi, K., Schelten, A., Silva,
  R., Smith, E.~M., Subramanian, R., Tan, X.~E., Tang, B., Taylor, R.,
  Williams, A., Kuan, J.~X., Xu, P., Yan, Z., Zarov, I., Zhang, Y., Fan, A.,
  Kambadur, M., Narang, S., Rodriguez, A., Stojnic, R., Edunov, S., and
  Scialom, T.
\newblock Llama 2: Open foundation and fine-tuned chat models, 2023.

\bibitem[Wang \& Manning(2010)Wang and Manning]{wang2010probabilistic}
Wang, M. and Manning, C.~D.
\newblock Probabilistic tree-edit models with structured latent variables for
  textual entailment and question answering.
\newblock In \emph{Proceedings of the 23rd International Conference on
  Computational Linguistics (Coling 2010)}, pp.\  1164--1172, 2010.

\bibitem[Wang et~al.(2023)Wang, Wei, Schuurmans, Le, Chi, Narang, Chowdhery,
  and Zhou]{wang2023selfconsistency}
Wang, X., Wei, J., Schuurmans, D., Le, Q., Chi, E., Narang, S., Chowdhery, A.,
  and Zhou, D.
\newblock Self-consistency improves chain of thought reasoning in language
  models, 2023.

\bibitem[Waterman \& Asanovi´c(2019)Waterman and
  Asanovi´c]{riscv-unpriviledged}
Waterman, A. and Asanovi´c, K.
\newblock The risc-v instruction set manual, volume i: User-level isa, document
  version 20191213.
\newblock 2019.
\newblock URL \url{https://riscv.org/technical/specifications/}.

\bibitem[Wei et~al.(2022)Wei, Bosma, Zhao, Guu, Yu, Lester, Du, Dai, and
  Le]{wei2022finetuned}
Wei, J., Bosma, M., Zhao, V.~Y., Guu, K., Yu, A.~W., Lester, B., Du, N., Dai,
  A.~M., and Le, Q.~V.
\newblock Finetuned language models are zero-shot learners, 2022.

\bibitem[Yao et~al.(2019)Yao, Ye, Li, Han, Lin, Liu, Liu, Huang, Zhou, and
  Sun]{yao-etal-2019-docred}
Yao, Y., Ye, D., Li, P., Han, X., Lin, Y., Liu, Z., Liu, Z., Huang, L., Zhou,
  J., and Sun, M.
\newblock {D}oc{RED}: A large-scale document-level relation extraction dataset.
\newblock In Korhonen, A., Traum, D., and M{\`a}rquez, L. (eds.),
  \emph{Proceedings of the 57th Annual Meeting of the Association for
  Computational Linguistics}, pp.\  764--777, Florence, Italy, July 2019.
  Association for Computational Linguistics.
\newblock \doi{10.18653/v1/P19-1074}.
\newblock URL \url{https://aclanthology.org/P19-1074}.

\end{thebibliography}
\bibliographystyle{icml2024}

%%%%%%%%%%%%%%%%%%%%%%%%%%%%%%%%%%%%%%%%%%%%%%%%%%%%%%%%%%%%%%%%%%%%%%%%%%%%%%%
%%%%%%%%%%%%%%%%%%%%%%%%%%%%%%%%%%%%%%%%%%%%%%%%%%%%%%%%%%%%%%%%%%%%%%%%%%%%%%%
% APPENDIX
%%%%%%%%%%%%%%%%%%%%%%%%%%%%%%%%%%%%%%%%%%%%%%%%%%%%%%%%%%%%%%%%%%%%%%%%%%%%%%%
%%%%%%%%%%%%%%%%%%%%%%%%%%%%%%%%%%%%%%%%%%%%%%%%%%%%%%%%%%%%%%%%%%%%%%%%%%%%%%%
\newpage
\appendix
%\onecolumn
%The $\mathtt{\backslash onecolumn}$ command above can be kept in place if you prefer a one-column appendix, or can be removed if you prefer a two-column appendix.  Apart from this possible change, the style (font size, spacing, margins, page numbering, etc.) should be kept the same as the main body.

\section{Entity extraction experiments}
\label{app01}

\begin{figure}[H]
	\centering
	\includesvg[inkscapelatex=false, width=3.2in]{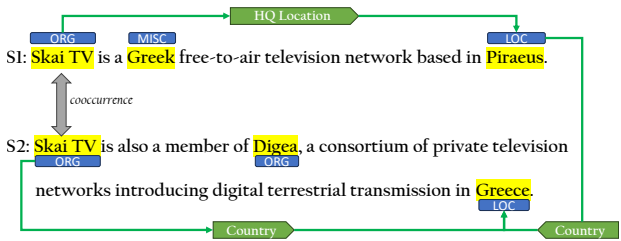}
	\caption{An example to illustrate what information is included in
		the DOCRED dataset. Each sentence is a separate data entry. 
		Relations and concurrences can cross multiple entries. \label{fig:docred_ex}}
\end{figure}

The experiments were based on the DOCRED dataset \cite{yao-etal-2019-docred}. 
The dataset is primarily for defining relations between so-called entities.
Hence, each sample includes human-annotated entities, their co-references, 
and triples showing the relation type between a pair of entities.
Figure \ref{fig:docred_ex} shows an example. Within each sentence, the 
entities are highlighted with their type (as blue). 
The relations are directional (as green). There is also a concurrrence 
between the subjects of the two sentences.

If one focuses only on the entity annotations and ignore the other two types, 
the dataset
can be utilized for studying the task of \textit{named entity recognition} (NER).
We use the dataset in our experiments in a similar way, by focusing
on the entity annotations. 
However, our focus is on {\it entity extraction}, not restricted to NER. 
Unlike NER which is a more well-defined problem
\cite{stanford-etextbook}, in our work we do not
attach a definition to ``entity extraction''. 
We simply let the oracle decide what it means and use the checker to
test which results are acceptable. 

Therefore, it is important to note that unlike a typical study of
a natural language task (e.g. NER) where the task is defined first
(e.g. by a benchmark dataset),
in our study the ``task'' itself (e.g. ``entity extraction'') 
is considered not well defined in advance. 
An oracle-checker implementation provides two perspectives to the
task, an oracle's view and a checker's view, separately. 
Between these two views, a result is decided to be accepted or not.
Hence, the accepted results reflect both views about the task. 
It would not be accurate to say that the task is defined by
the checker alone or the oracle alone.

The oracle-checker way to define a task can become handy when
dealing with a domain-specific input where the task is specific
to the domain expert. For example, a hardware design company 
might desire to perform ``entity extraction'' on their in-house
design documents. These documents are specific and confidential
to the company. When they think about ``entity extraction'', they
might have a different idea about what ``entities'' mean, 
which is different from the standard NER practice. 
Instead of preparing their own domain-specific dataset to train
their own entity extraction model, as an alternative, they can
implement their ideas of entity acceptability into a checker and
use the oracle-checker scheme to ``filter-in'' the entities
they are looking for. 

In view of such an application context, in our experiments we also
used the RISC-V unprivileged ISA specification \cite{riscv-unpriviledged} 
as a separate data source. The RISC-V specification has no 
entity annotations. Our interest was to see if ``entity extraction''
on such a document would behave differently from that on the
DOCRED sentences. As seen in section~\ref{sec05.01}, the behavior on
the RISC-V specification is quite different. 
The linearity test checker allows us to characterize this difference
from a particular perspective, in this case asking whether the entity
extraction is linear or not. 

\subsection{Prompts}
\label{app.entity.prompts}

In our experiments, we did not focus on optimizing the prompts to
GPT3.5. Prompt engineering was only pursued to the point of 
getting {\it consistent} and {\it valid} responses. 
To follow common convention in NER, our prompt also asks GPT3.5 
to provide a {\it class} (or type) for each entity. 
Classes in the DOCRED dataset includes such as 
``people'', ``location'', ``organization'', and so on.
As a contrasting example, 
the chemical NER dataset CHEMDNER \cite{Krallinger2015}
(which we did not use), 
includes classes of ``abbreviation'', ``family'', and ``formula'' (of chemicals),
and so on. 
It might typically expected that an entity is formed only by consecutive words from the text. 
However, we did not enforce this constraint. For example, consider that for the 
phrase ``William and Annie Washington". GPT3.5 would extract both ``William Washington''
and ``Annie Washington'' as entities. 
In contrast, the DOCRED labels only have ``William''
for the first entity.

Prompts are used for two purposes in our linearity checking experiments: to perform
the entity extraction task and to help implement the synonym generator
$syn$ discussed in section~\ref{sec03.1}.

Prompt~\ref{app.prompt.entity} shows the specific prompt for entity extraction.
We gave an example entity list as the output in the prompt, 
but we did not provide an example input nor explicitly define classes.
This is because we did not want to over-constrain the responses, 
even though adding a full in-context example might 
increase the number of valid responses.

\setcounter{algorithm}{0}
\begin{algorithm}[h]
	\floatname{algorithm}{Prompt}
	\caption{Entity extraction}
	\label{app.prompt.entity}
	\scriptsize\ttfamily
	Please perform entity recognition for all entities in the following sentence: "\{text\}"\\
	\\
	Present the result as a strictly formatted numbered list e.g.\\
	"1. movie: Wizard of Oz\\
	2. animal: tiger"\\
\end{algorithm}

Prompts \ref{prompt:named}, \ref{prompt:replace}, and \ref{prompt:syn} are used in the synonym generator. 
Prompt \ref{prompt:named} is a decision-making prompt to determine if an extracted entity is name based. 
%We manually evaluated the decisions over 100 sentences, leading us to include a  short chain-of-thought to the prompt \cite{wei2023chainofthought} to improve the decisions. 
Prompts \ref{prompt:replace} and \ref{prompt:syn} are generation prompts, and 
produce the actual text to replace entities in our sentence transformations. 
For these prompts we include a suggestion for formatting.

% \vspace{-3mm} % Adjust the value as needed
\begin{algorithm}[H]
	\floatname{algorithm}{Prompt}
	\caption{Named/non-named classification}
	\label{prompt:named}
	\scriptsize\ttfamily
	Is the entity "\{entity\}" in the sentence "\{text\}" a "named entity" or a "normal entity"?
	Please explain in 20 words or less, and then place your answer in double brackets [[ ]]
\end{algorithm}
% \vspace{-3mm} % Adjust the value as needed
\begin{algorithm}[H]
	\floatname{algorithm}{Prompt}
	\caption{Replacement of named entity}
	\label{prompt:replace}
	\scriptsize\ttfamily
	Please list five random entities (type: "\{entity\_type\}") that could replace "\{entity\}" in the sentence "\{text\}".\\
	
	Format your answer with a numbered list of the synonyms. e.g.\\
	"1. Synonym1\\
	2. Synonym2"\\
\end{algorithm}
% \vspace{-3mm} % Adjust the value as needed
\begin{algorithm}[H]
	\floatname{algorithm}{Prompt}
	\caption{Synonym list of non-named entity}
	\label{prompt:syn}
	\scriptsize\ttfamily
	Please list synonyms for "\{entity\}" in the sentence "\{text\}".\\
	
	Format your answer with a numbered list of the synonyms. e.g.\\
	"1. Synonym1\\
	2. Synonym2"\\
\end{algorithm}

\subsection{Detail of the synonym generator}
\label{app.entity.synonyms}

% \subsubsection*{Named or not? Synonym Rules}
The synonym generator $syn$ was divided into two parts: (1) decide if an entity is named, and (2) assign an entity replacement or synonym. Initially, GPT3.5 by itself 
was considered as a convenient candidate for $syn$, but in an effort to reduce calls and uncertainty we also implemented a number of intuitive rules to facilitate synonym generation.

\textbf{1. Named/non-named classification.}
Every extracted entity was passed through a series of capitalization based rules as a baseline 
to decide if it is named or not-named. 
Only cases that did not match these rules were passed to GPT3.5 for the
named/non-named decision.
There were also some universal rules 
(e.g. \textit{lowercase and non-numeric = non-named}), 
and class-based rules that were chosen based on 
GPT3.5's most common class labels (\textit{person, location, organization, date, film, video game\dots}).
For example, a \textit{person} with any capitalization is named (``Aaron'' vs ''my teacher''). 
Only 7\% of GPT extracted entities in the 5000 DOCRED sentences could not be decided by the rules. 
For those, prompt \ref{prompt:named} is used with a self-consistency vote out of 11 to make 
the final named/non-named decision.

\textbf{2. Synonym generation.} 
If a given entity is named, it is first processed by additional custom rules. 
For example, named entities belonging to a commonly extracted class, like ``film'', 
were replaced by selecting from the collective pool of all ``film'' entities.
For entities not matching our rules, 
then GPT was asked to {\it replace} them with another entity (prompt \ref{prompt:replace}). 
Non-named entities went straight to GPT asking for a synonym list (prompt \ref{prompt:syn}). 
For both prompts, the final synonym selection was based on a self-consistency vote over 7 repeated
runs. 
%(each run would yield multiple synonym options, so the lists were aggregated before voting). 

Prompt \ref{prompt:replace} and prompt \ref{prompt:syn} processes named and non-named
entities separately. To give an example why this is required,
consider the synonyms of Barack Obama and of John Smith.  
According to GPT3.5, candidates 
for Barack Obama include [``the 44th president of the United States", ``Former president Barack Obama"].
Candidates for John Smith include [``he'', ``Johnny Smith'', ``Mr.Smith'']. 
For example, it might be unlikely that ``the 44th president of the United States" 
and ``he" to be extracted as a single entity, 
reducing the chance of passing the linearity test. 
To exclude those special cases, 
in our current implementation 
named entities are simply replaced by another name of the same type,
e.g. another person, another date, and so on.
This is to ensure that a replaced entity has little chance to share names 
with the original entity.  

\begin{figure*}[htb]
	\centering
	\fbox{
		\includesvg[inkscapelatex=false, width=5in]{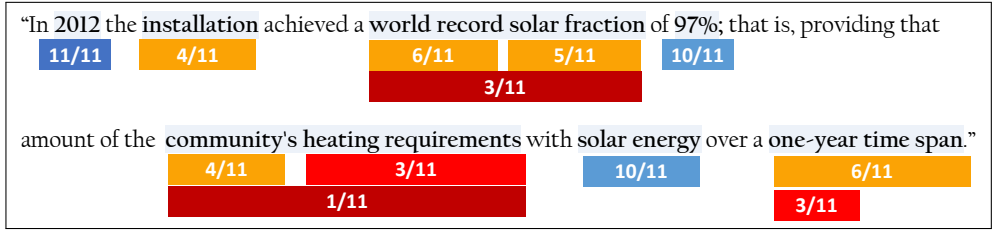}
	}
	\caption{A multi-form entity example: 
		The sentence is rejected by the test. 
		Among the 11 repeated runs, the number of times the entity appears in
		the result is shown under the entity. For example, the ``11/11'' under the 
		entity ``2012'' means the entity appears 11 times during 11 runs. 
		As seen, the entities ``world record'' and ``solar fraction'' together can form
		a longer entity. This is similar for ``community'', and ``heating requirements''. 
		For example, if an entity do not appear most of the times (e.g. $>$ 5/11), they can be
		considered as weak. 
		Then there are many weak entities in the example. 
		Note that this definition of being weak is subjective. 
		\label{fig:failcasemulti}}
\end{figure*}

\begin{figure*}[htb]
	\centering
	\fbox{
		\includesvg[inkscapelatex=false, width=5.5in]{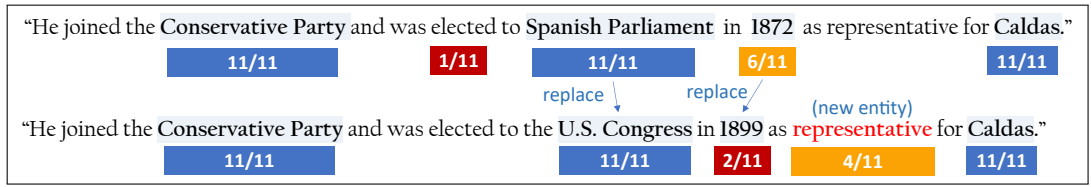}
	}
	\caption{A weak entity example (a rejected sentence): 
		The first sentence is transformed into the second sentence in the linearity test
		where two entities are replaced with their synonym. 
		Each entity is marked with its consistency. 
		The entity ``representative'' appears with the transformed sentence after
		the two synonyms replacing its original entity. 
		In the new sentence, the entity ``1899'' becomes weaker 
		and the entity ``elected'' never appears. 
		The result indicates three weak entities, ``elected'', ``1899'', and
		``representative''. 
	\label{fig:failcaseweak}}
\end{figure*}

\subsection{Examples of rejected entities}
\label{app.entity.fails}

A by-product of our linearity test is a collection of those rejected cases.
In our experiments, we attempted to categorize them to gain more insight
on the entity extraction behavior of the oracle.
The findings shown in section~\ref{sec05.01} include two aspects:
(1) rejection tends to correlate to self-consistency, and (2) the behavior on
a domain-specific document like RISC-V specification tends to be
different from that on a public dataset like DOCRED. 
In this part of the Appendix, we provide some hypotheses behind
these findings. 

We observed two types of rejection behaviors that might contribute to reduced consistency.
We refer to them as \textbf{weak} entities and \textbf{multi-form} entities. 
These two behaviors can overlap. 
Note that in section~\ref{sec05.01}, we mentioned that we considered any of the 
$11^3$ combinations passing the test as passing. Hence, from a binary pass/fail
point of view, a failing entity might eventually become passing if we keep
increasing the number of repeated runs. Hence, the discussion for the ``failing
cases'' below should be seen in view of this limitation. 
 
In addition to weak and multi-form entities, a third failure type exhibits 
behavior similar to multi-form, but manifests in the results of the 
synonym generator $syn$ rather than the entity extraction itself.
A fourth failure type, called {\it assumed knowledge}
reflects that there are some entities whose extraction {\it does} depend upon 
the existence of others. This would be an obvious situation where the
entity extraction becomes ``non-linear'' from our checker's point of view. 
It is also interesting to mention another observation: If an entity is
a named entity, though still possible, 
it is less likely that it will cause the first 
two types of failures,  

\begin{table}[htb] % 'h' here is the placement specifier, meaning 'here'
	\centering
	\caption{
		Examples of synonyms that, if not dealt with, can cause a failure in the linear test.
		(1) Replacing ``Roald" with its proposed synonym effectively repeats the entire sentence, artificially adding multiple entities.
		(2) The entity ``Wallonia" has a suggested replacement of ``Liege, Belgium", but ``Belgium'' was already an entity in the original sentence.
		(3) GPT3.5 considers ``grant'' as a possible replacement for ``US\$ 1.826 billion'', 
		resulting in a phrase ``...was worth grant'' which exhibits weak behavior for ``grant''.
		\label{app.table.synproblems}}
	\textit{``Vigra Church is located a short distance south of Roald.''}
	\begin{tabular}{cc}
		\toprule
		\textbf{Entity} & \textbf{GPT3.5 Synonym}\\
		\midrule
		Vigra Church & Stavanger Cathedral \\
		Roald & Vigra Church ... south of Ålesund. \\
		\bottomrule
	\end{tabular}
	\vspace{0.2in}
	
	\textit{``Crupet is a village in Wallonia, Belgium.''}
	\begin{tabular}{cc}
		\toprule
		\textbf{Entity} & \textbf{GPT3.5 Synonym}\\
		\midrule
		Crupet & Charleroi \\
		Wallonia & Liege, Belgium \\
		Belgium & France\\
		\bottomrule
	\end{tabular}
	\vspace{0.2in}
	
	\textit{``The award, along with funds for the construction of, was worth US\$ 1.826 billion.''}
	\begin{tabular}{cc}
		\toprule
		\textbf{Entity} & \textbf{GPT3.5 Synonym}\\
		\midrule
		construction & building \\
		US\$ 1.826 billion & grant \\
		\bottomrule
	\end{tabular}
\end{table}
\begin{table}[h]
	\centering
	\caption{
		An example due to GPT3.5's own assumed knowledge: 
		``Twilight'' is extracted as an entity, 
		due to its knowledge about the novel associated 
		with the characters Bella, Edward, and Jacob. \label{twilight}
	}
	\begin{tabular}{p{75mm}}
		\toprule
		% sentence 499, 2 in docred 
		\textit{``The novel explores {\color{blue}Bella}'s compromise between her love for {\color{blue}Edward} and her friendship with shape-shifter {\color{blue}Jacob Black}, along with her dilemma of leaving her mortality behind in a terrorized atmosphere, a result of mysterious vampire attacks in {\color{blue}Seattle.}''} \\
		$\longrightarrow$ [{\color{blue}Bella, Edward, Jacob Blake, Seattle, \textbf{Twilight}}]\\
		\midrule
		\textit{``The novel explores {\color{red}B'Elanna's} compromise between her love for {\color{red}Epidicus} and her friendship with shape-shifter {\color{red}Jack Ryan}, along with her dilemma of leaving her mortality behind in a terrorized atmosphere, a result of mysterious vampire attacks in {\color{red}San Fransisco}''}\\
		$\longrightarrow$ [{\color{red} B'Elanna, Epidicus, Jack Ryan, San Fransisco}]\\
		\bottomrule
	\end{tabular}
\end{table}

\subsubsection*{Type 1: ``Multi-form entities''}
\label{app.entity.failcases.1}

A multi-form entity is an entity that can be extracted in a variety of ways.
Figure~\ref{fig:failcasemulti} shows an example.
Named entities, when causing a test to reject, 
were more susceptible to multi-form behavior than weak behavior. 
For example, ``Mohammed's family'' could become multi-form entities, 
with GPT either extracting the full possessive, the person (``Mohammed''), or both. 
Chained named entities, like ``Wallonia, Belgium'' or ``January of 2010'' 
could also be extracted as separate entities or all together.

\subsubsection*{Type 2: ``Weak'' entities}

An entity might be considered as weak when it rarely, or  
inconsistently appears across repeated entity extraction runs. 
We have observed cases where a weak entity could be caused by 
inter-dependency with another entity.
Most observed cases seemed to stem from situations where 
selecting those weak entities could be highly subjective.

Figure \ref{fig:failcaseweak} shows an example with weak entities. 
It is interesting to observe when ``Spanish Parliament'' and ``1872'' 
in the original sentence are replaced with ``U.S. Congress'' and ``1899'',
the new entity ``representative'' is extracted, 
indicating it is a weak entity in the original sentence. 
%From our observation, dates, unusually for named entities, often exhibit as weak,
%e.g. a year ``1632'', a period ``early 1890s'', or an exact day ``July 15, 1996''.

In an application of the linearity test, determining the cause
for a weak entity might 
not be as crucial. The test ensures that the accepted entities are all ``strong''
in the sense that they pass the test, i.e. the test is used to
filter in ``strong'' entities, rather than filtering out specific weak entities. 

% % Figure \ref{fig:failcaseweak} shows the high variation of a more difficult sentence. 
% % GPT extraction on this sentence resulted in 11 unique entity sets (consistency=1).

% % Note - as mentioned before, failure entities can exhibit multiple behaviors. Example (1b) does have entities with multi-form behavior too, like ``loveless partnership".

% The named entities that were observed to often portray the \textit{weak} quality were dates - . 

\subsubsection*{Type 3: Issues of synonym generation}

Occasional issues in the synonym generation,
if not dealt with, 
could affect the linearity test result. 
Table~\ref{app.table.synproblems} shows three such examples. 
This is to show that optimization of the synonym generation
can improve the acceptance rates. 
We leave further optimization to future work.

\subsubsection*{Type 4: Assumed Knowledge}
Sometimes variations in entity extraction could be due to 
GPT3.5 assumed knowledge about the sentence, 
such as abbreviation expansions (in RISC-V, ``CSR'' could be 
extracted as ``control status register'') or by inferring a new entity that
does not appear in the sentence. 
Table~\ref{twilight} shows an example of this type.
It is interesting to see that GPT3.5, due to its own knowledge
about the characters mentioned in the sentence, extracted the
entity ``Twilight'' which is associated with the novel. 
After the characters and place name are replaced, 
the entity no longer appears.

\subsection{On the difficulty of entity extraction for the RISC-V specification}
\label{app.entity.riscv}

As shown in Figure~\ref{linearity}, sentences from the RISC-V specification 
have a much higher rejection rate than those from the DOCRED dataset. 
One possible explanation is that, most extracted
entities from the RISC-V specification are not named entities. 
Results from our synonym generation indicated that for DOCRED sentences,
roughly 76\% of the extracted entities were named.
This percentage dropped to 36\% for sentences from the RISC-V specification. 

As mentioned above, named entities tend to be ``more stable'' than other entities.
A public dataset like DOCRED is often used for study of NER. 
As a result, entities extracted from the dataset are more stable from
the weak and multi-form perspectives. 
Non-named entities, on the other hand, can be ambiguous even to human eyes and 
can also be context-sensitive. 
As an example, the following shows a sentence from the RISC-V specification.

\begin{quote}
\it
RV32E reduces the integer register count to 16 general-purpose registers, (x0–x15), where x0 is a dedicated zero register.
\end{quote}

The term ``RV32E'' can be a named entity because it is a well-defined term in
the architecture specification. 
However, there can be multiple potential entities from 
the phrase \textit{``16 general-purpose registers''}.
The following table lists four of the candidates to consider. 
Each candidate might be justifiable with its own reason.
Depending on a particular reason chosen by the oracle during entity extraction, 
a particular entity might be selected. 
This can be seen as an example of the multi-form type discussed before. 

\begin{table}[H]
	
	\centering
	\small
	\begin{tabular}{p{38mm}p{35mm}}
		\toprule
		Variation &  Justification \\\midrule
		16 & NER often considers numbers as entities\\\midrule
		16 general-purpose registers & a count of anything can be an entity\\\midrule
		general-purpose registers & specific type of register \\\midrule
		registers & a general reference\\
		\bottomrule 
	\end{tabular}
	\caption{RISCV ``16 general-purpose registers'' entity variations}
	\label{app.table.riscvmultiform}
\end{table}

\subsection{Detailed results on the correlations}
\label{app.entity.corr}

\begin{figure*}[htb]
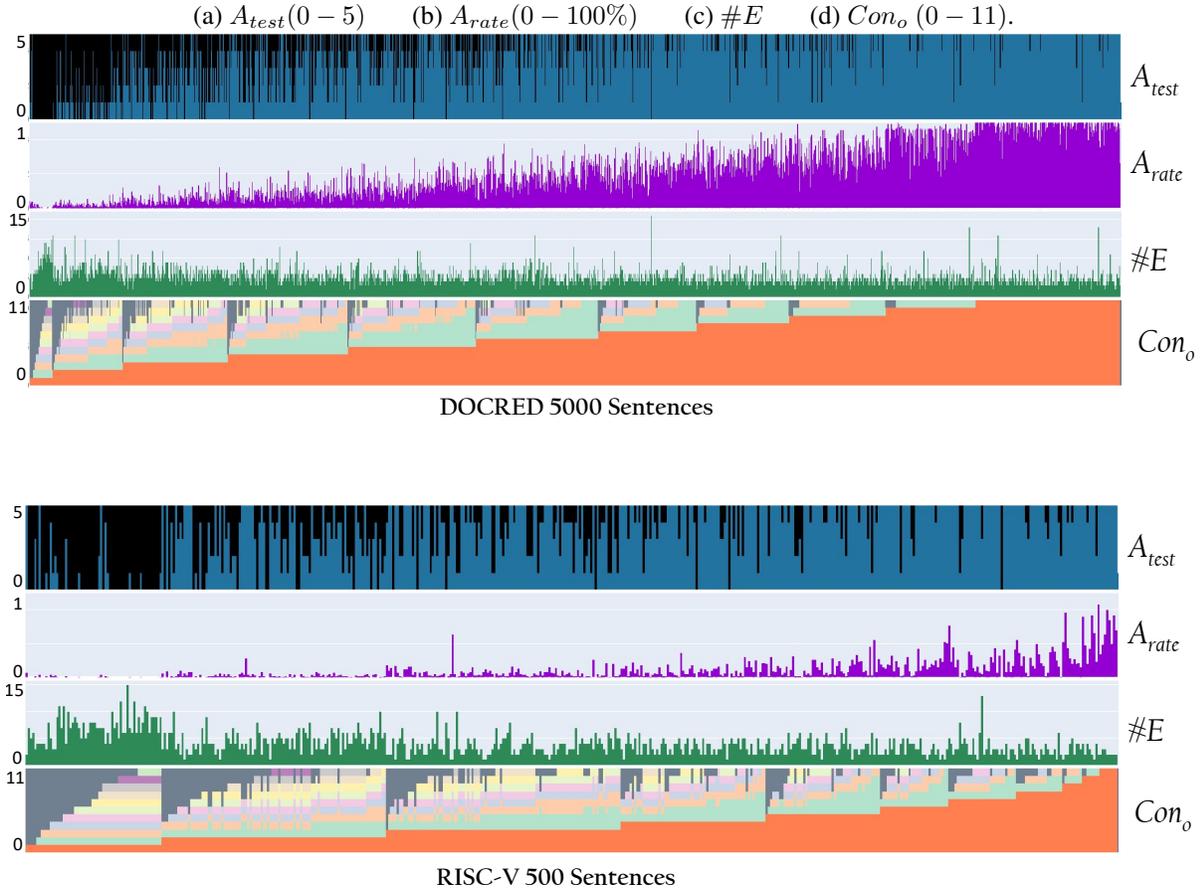

	\centering
	(a) $A_{test} (0-5)$ \ \ \ \ \ \ 
	(b) $A_{rate} (0-100\%)$  \ \ \ \ \ \ 
	(c) $\#E$ \ \ \ \ \ \ 
	(d) $Con_o \ (0-11)$.
	\includesvg[inkscapelatex=false, width=6.2in]{Figures/svg/thomas2.svg}
	\par \vspace{10mm}
	\includesvg[inkscapelatex=false, width=6.2in]{Figures/svg/thomas3.svg}
	\caption{For each of the terms, $A_{test}$, $A_{rate}$, \#$E$, and $Con_o$,
		defined with Table~\ref{correlations}, a bar chart shows its
		actual values across the sentences. 
		For the $A_{test}$ bar charts, color blue indicates
		that the test passes. 
		For $A_{rate}$, color purple indicates the passing rate.
		For \#$E$, color green indicates the number of entities.
		For $Con_o$, color orange indicates the size of the most
		self-consistent group. Other colors indicate the sizes
		of their respective groups. A dark gray indicates not
		being able to get a valid result.  
		Both plots are sorted based $Con_o$. 	\label{app.fig.correlation}
	}

\end{figure*}

Table~\ref{correlations} shows the correlation results. 
Figure~\ref{app.fig.correlation} plots the actual values of the
four terms defined in the correlation table:
$A_{test}$, $A_{rate}$, \#$E$, and $Con_o$.
Two plots are shown, one for the DOCRED sentences and the other
for the sentences from the RISC-V specification. 
Both plots are sorted based on the values of $Con_o$.

If we focus on $A_{rate}$, we can see that the passing
rates of DOCRED sentences tend to be much higher
than RISC-V sentences. 
The correlation between $A_{rate}$ and $Con_o$ can be
observed with their respective bar charts, in both 
DOCRED and RISC-V cases. 
Note that while in retrospect, the correlation
between $A_{rate}$ and $Con_o$ might be seen as somewhat 
intuitive, it was not anticipated in advance and the 
relationship was observed only after running the 
checker. 

\newpage

\section{Paraphrase decision experiments}
\label{app02}

Each sample in the MSRP dataset is represented as
two sentences $s_1, s_2$ and a label indicating if
they are semantically equivalent. 
In this part of the Appendix, we use examples to further
illustrate the results presented in section~\ref{sec05.02}. 

\subsection{Prompts}
\label{app02.1}

The experiments involve three prompts. 
Prompt~\ref{paraphrase decision} shows the main prompt which queries
GPT3.5 for the semantic equivalence on a pair of sentences. 
The ``System Message'' and ``Human Message'' are not part of the
prompt. They are for indicating the different portions of the prompt. 

\begin{algorithm}[h]
	\floatname{algorithm}{Prompt}
	\caption{Query for semantic equivalence: yes or no \label{paraphrase decision}}
	\scriptsize
	System Message: \\
	{\ttfamily
		"You are a helpful assistant that decides if two sentences are paraphrases designed to output JSON."}
	Human Message: \\
	{\ttfamily
		"Decide whether the following two sentences are semantically equivalent: \\
		1. '\{sentence1\}' 2. '\{sentence2\}' \\
		Answer yes or no and provide a short explanation. Output with keys 'answer' and 'explanation'. "}
\end{algorithm}

Like that in the linearity test experiments, we did not focus on
optimization of the prompts. Prompt engineering was only
pursued to the point of enabling meaningful experiments to
be conducted. 
It is also interesting to point out that the prompts were fixed,
long before our checker designs and implementations were completed. 

As discussed with the Algorithm~\ref{alg:search} before, 
the function $align()$ involves asking the oracle whether two
matching phrases are semantically equivalent or not.
Prompt~\ref{phrase alignment decision} shows the prompt used
for this purpose. The ``text1'' and ``text2'' in the prompt
refer to two matching phrases in a given potential alignment
found by the algorithm.  

\begin{algorithm}[h]
	\floatname{algorithm}{Prompt}
	\caption{Query for phrase equivalence: yes or no \label{phrase alignment decision}}
	\scriptsize 
	System Message: \\
	{\ttfamily
		"You are a helpful assistant that decides if two texts could be paraphrases designed to output JSON."} \\
	Human Message: \\
	{\ttfamily
		"Decide whether the following two texts could be considered semantically equivalent: \\
		1. '\{text1\}' 2. '\{text2\}' \\ 
		Answer yes or no and provide a short explanation. Output with keys 'answer' and 'explanation'. "}
\end{algorithm}

\begin{algorithm}[h]
	\floatname{algorithm}{Prompt}
	\caption{Query for generating paraphrases \label{paraphrase generation}}
	\scriptsize 
	System Message: \\
	{\ttfamily
	"You are a helpful paraphrases generator designed to output JSON."} \\
	Human Message: \\
	{\ttfamily
	"Generate five paraphrases that are semantically equivalent to the following sentence: '\{sentence\}'\\
	Output with key 'paraphrases'."}
\end{algorithm}

As mentioned before with Figure~\ref{nonisomorphism}, the $par$ function 
is implemented by asking the oracle to generate a set of paraphrases
for the given sentence. Prompt~\ref{paraphrase generation} shows the prompt
for this purpose. 
%Note that instead of generating one paraphrase after another, the prompt asks for five paraphrase at once, to speed up the run time. 

\begin{table*}[ht]
	\centering
	\caption{Sentence pairs ($s_1, s_2$) from MSRP dataset where GPT3.5's decision on semantic equivalence is different from annotated labels. In each case, GPT3.5 says they are semantically 
		equivalent and the labels say the opposite, and
		GPT3.5 provides an explanation for its decision. Major differences in 
		wording are highlighted in color. \label{tab:AppB_disagreement1-3}}
	\begin{tabular}{p{0.4\textwidth}p{0.4\textwidth}cc}
		\toprule
		$s_1$ & $s_2$ & Label & GPT3.5 \\
		\midrule
		% 1
		Legislation making it harder for consumers to erase their debts in bankruptcy court won overwhelming House approval in March. & 
		Legislation making it harder for consumers to erase their debts in bankruptcy court won speedy, House approval in March \textcolor{blue}{and was endorsed by the White House.} &  
		No & Yes \\
		\multicolumn{4}{p\textwidth}{\it {\bf GPT3.5 Explanation:} Both sentences convey the same information about legislation that makes it harder for consumers to erase their debts in bankruptcy court, winning approval in the House in March and being endorsed by the White House.} \\ \midrule
		% 2
		\textcolor{orange}{As well as the dolphin scheme,} the chaos has allowed foreign companies to engage in damaging logging and fishing operations without proper monitoring or export controls. &
		Internal chaos has allowed foreign companies to set up damaging commercial logging and fishing operations without proper monitoring or export controls. &
		No & Yes \\  
		\multicolumn{4}{p\textwidth}{\it {\bf GPT3.5 Explanation:} Both sentences convey the idea that internal chaos has enabled foreign companies to conduct damaging logging and fishing operations without proper oversight or controls.} \\ \midrule
		% 3
		"We're just dealing with bragging rights here, who wins and who loses." &  "\textcolor{blue}{Leaving aside attorney fees,} we're dealing with bragging rights of who wins and who loses," \textcolor{blue}{said Gammerman.} & 
		No & Yes \\
		\multicolumn{4}{p\textwidth}{\it {\bf GPT3.5 Explanation:} Both sentences convey the same meaning that the issue at hand is about bragging rights and determining who wins and who loses, with the mention of attorney fees being the only difference.} \\ \bottomrule
	\end{tabular}
\end{table*}

\begin{table*}[ht]
	\centering
	\caption{Similar to Table~\ref{tab:AppB_disagreement1-3}, in these three cases,
		GPT3.5 says they are not semantically equivalent and the labels say they are. \label{tab:AppB_disagreement5-7}}
	\begin{tabular}{p{0.4\textwidth}p{0.4\textwidth}cc}
		\toprule
		$s_1$ & $s_2$ & Label & GPT3.5 \\
		\midrule
		% 5
		\textcolor{orange}{They} had published an advertisement on the Internet on June 10, offering the \textcolor{orange}{cargo} for sale, he added. &  
		On June 10, \textcolor{blue}{the ship's owners} had published an advertisement on the Internet, offering the \textcolor{blue}{explosives} for sale. &  
		Yes & No \\
		\multicolumn{4}{p\textwidth}{\it {\bf GPT3.5 Explanation:} The two sentences are not semantically equivalent. The first sentence mentions 'cargo' being offered for sale, while the second sentence mentions 'explosives' being offered for sale. Additionally, the first sentence does not mention the ship's owners, while the second sentence does.} \\ \midrule
		% 6
		\textcolor{orange}{The new Finder} puts a user's folders, hard drive, network servers, iDisk and removable media in one location, \textcolor{orange}{providing one-click access.} &  
		\textcolor{blue}{Panther's redesigned Finder navigation tool} puts a user's \textcolor{blue}{favourite} folders, hard drive, network servers, iDisk and removable media in one location. &  
		Yes & No \\
		\multicolumn{4}{p\textwidth}{\it {\bf GPT3.5 Explanation:} The two sentences are not semantically equivalent. The first sentence mentions 'one-click access' while the second sentence mentions 'favorite folders' and 'redesigned Finder navigation tool', indicating a difference in functionality and design.} \\ \midrule
		% 7
		The company has said it plans to restate its earnings for \textcolor{orange}{2000 through 2002.} &  
		The company had announced \textcolor{blue}{in January} that it would have to restate earnings for \textcolor{blue}{2002, 2001 and perhaps 2000.} &  
		Yes & No \\
		\multicolumn{4}{p\textwidth}{\it {\bf GPT3.5 Explanation:} The first sentence mentions restating earnings for 2000 through 2002, while the second sentence specifies restating earnings for 2002, 2001, and perhaps 2000. The differences in the years mentioned make the sentences not equivalent.} \\
		\bottomrule
	\end{tabular}
\end{table*}

\begin{table*}[ht]
	\centering
	\caption{Similar to Table~\ref{tab:AppB_disagreement1-3}, in this case,
		the differences appear on both sentences. \label{tab:AppB_disagreement4}}
	\begin{tabular}{p{0.4\textwidth}p{0.4\textwidth}cc}
		\toprule
		$s_1$ & $s_2$ & Label & GPT3.5 \\
		\midrule
		% 4
		The Justice Department filed suit Thursday against the state of Mississippi \textcolor{orange}{for failing to end what federal officials call "disturbing" abuse of juveniles and "unconscionable" conditions} at two state-run facilities. &
		The Justice Department filed a civil rights lawsuit Thursday against the state of Mississippi, \textcolor{blue}{alleging abuse of juvenile offenders} at two state-run facilities. &  
		No & Yes \\ 
		\multicolumn{4}{p\textwidth}{\it {\bf GPT3.5 Explanation:} Both sentences convey the same information about the Justice Department filing a lawsuit against the state of Mississippi for the mistreatment of juvenile offenders at two state-run facilities.} \\ \bottomrule
	\end{tabular}
\end{table*}

\subsection{Label disagreements by GPT3.5}
\label{app02.2}
Following the result shown in Table~\ref{MSRPvsGPT}, Table~\ref{tab:AppB_disagreement1-3} 
shows three cases where GPT3.5's ``Yes'' decisions disagree with the annotated ``No'' labels. 
The differences between the three pairs of sentences are highlighted in color.
It is interesting to observe that in all three cases, one sentence simply
adds ``more information'' to the other, i.e. the colored phrase appears in
one but not the other. GPT3.5's explanations for their semantic equivalence
basically consider the two sentences ``convey the same information'',
``convey the (same) idea'', and ``convey the same meaning'', respectively. 
This indicates that GPT3.5 does not think the additional information on
one sentence alters the main information content which is the same between the two
sentences. 

Intuitively, the three examples might be seen as providing 
some support for our implementation
of the $align(U, S_U)$ function in Algorithm~\ref{alg:search}. As stated
then, the $align()$ function includes checking if all words (except for
propositional words) in the source sentence
are covered by matching phrases in $S_U$.
If we consider finding a $\rho$-alignment for the three examples,
the $align()$ is designed in such a way that we desire to see
them aligned as
$\rho_{s_1 \rightarrow s_2}$ ($s_1$ is covered),
$\rho_{s_2 \rightarrow s_1}$ ($s_2$ is covered), and
$\rho_{s_1 \rightarrow s_2}$ ($s_1$ is covered), respectively.
Based on the examples, we observe that 
our ``Yes'' prover, which is based on $align()$, 
is more compatible with GPT3.5 when making a decision on
semantic equivalence. 

However, it should be emphasized that
we did not design the ``Yes'' prover to approximate how GPT3.5
decides on semantic equivalence nor how it makes a judgment call
when doing so. As mentioned in section~\ref{sec04.1},
our ``Yes'' prover was inspired by the idea of compositional
phrase alignment \cite{arase-tsujii-2020-compositional}\cite{wang2010probabilistic},
independently of GPT3.5. The connection shown in the three examples
was only observed during the experiments. 

In contrast, Table~\ref{tab:AppB_disagreement5-7} shows
three examples where GPT3.5 says they are not semantically 
equivalent and the labels say they are. 
It is interesting to see that in these examples, the differences
appear on both sides. The differences are also  
mentioned in GPT3.5 explanations. 

It should be noted that whether the differences appear on one side
or on two side, is {\it not} the only way GPT3.5 makes its decision.
Table~\ref{tab:AppB_disagreement4} shows an example where the 
differences appear on both sides and GPT3.5 says they are semantically
equivalent as well. 
In a sense, GPT3.5 considers the differences
semantically equivalent and hence, the two sentences 
``convey the same information''. 

For all the seven cases above, one might be able to argue 
for the labels to be correct. This depends on one's perspective on
the definition of semantic equivalence. 
These examples illustrate how deciding on
semantic equivalence can involve various judgment calls, 
as pointed out at the beginning of section~\ref{sec05.02} before. 

As shown in our sanity check in section~\ref{sec05.02.0}, 
when the two sentences are obviously not semantically equivalent,
GPT3.5's decisions agree with our checker on all. 
If two sentences are (almost) identical, then there is little
need to make a judgment call. A judgment call is
needed when the semantic equivalence of 
two sentences is not that obvious. 

\subsection{Examples from the ``Yes'' cases}
\label{app02.3}

From Table~\ref{MSRPvsGPTvsChecker}, for those GPT3.5 ``Yes'' cases,
we see that our checker can reach two outcomes:
some rejected by the checker indicating failure for finding a proof
and others accepted because of a found proof. 

\begin{table*}[ht]
	\centering
	\caption{Three examples of leftover words after alignment, 
		which are ``Yes'' prover rejected cases. \label{tab:AppB_rejected}}
	\begin{tabular}{p{0.4\textwidth}p{0.4\textwidth}}
		\toprule
		$s_1$ & $s_2$ \\
		\midrule
%		(This is wrong)
%		The DVD-CCA then appealed to the state Supreme Court. &
%		The DVD CCA appealed that decision to the U.S. Supreme Court.\\
%		\multicolumn{2}{l}{\it {\bf Post-alignment leftover: } $s_1$: \{ \textcolor{orange}{then, state} \}, $s_2$: \{ \textcolor{blue}{that decision, U.S.} \} } \\
%		\midrule 
		But he added group performance would improve in the second half of the year and beyond. &
		De Sole said in the results statement that group performance would improve in the second half of the year and beyond. \\
		\multicolumn{2}{l}{\it {\bf Post-alignment leftover: } $s_1$: \{ \textcolor{orange}{But he} \}, $s_2$: \{ \textcolor{blue}{De Sole} \} } \\
		\midrule 
		The first biotechnology treatment for asthma, the constriction of the airways that affects millions around the world, received approval from the US Food and Drug Administration yesterday. &
		The first biotechnology treatment for asthma, the constriction of the airways that affects millions of Americans, received approval from the U.S. Food and Drug Administration on Friday. \\
		\multicolumn{2}{l}{\it {\bf Post-alignment leftover: } 
			$s_1$: \{ \textcolor{orange}{around the world} \}, 
			$s_2$: \{ \textcolor{blue}{Americans} \} } \\
		\midrule 
%		(This is wrong)
%		Gyorgy Heizler, head of the local disaster unit, said the coach was carrying 38 passengers. &
%		The head of the local disaster unit, Gyorgy Heizler, said the coach driver had failed to heed red stop lights.\\
%		\multicolumn{2}{l}{\it {\bf Post-alignment leftover: } $s_1$: \{ \textcolor{orange}{was carrying 38 passengers} \},} \\
%		\multicolumn{2}{l}{\hspace{1.4in} $s_2$: \{\textcolor{blue}{driver had failed to heed red stop lights} \} } \\
%		\midrule
		The University of Michigan released a new undergraduate admission process Thursday, dropping a point system the U.S. Supreme Court found unconstitutional in June. &
		The University of Michigan released today a new admissions policy after the U.S. Supreme Court struck down in June the way it previously admitted undergraduates. \\
		\multicolumn{2}{l}{\it {\bf Post-alignment leftover: } $s_1$: 
			\{ \textcolor{orange}{Thursdays, dropping a point system, found unconstitutional} \}, } \\
		\multicolumn{2}{l}{\hspace{1.4in} 
			$s_2$: \{\textcolor{blue}{today, struck down, way it previously admitted undergraduates} \} } \\
		\bottomrule
	\end{tabular}
\end{table*}

\subsubsection{Examples rejected by the checker}

Table~\ref{tab:AppB_rejected} shows three examples to illustrate what failures
mean in our ``Yes'' prover. A failure means that after the attempt of
$align()$ in Algorithm~\ref{alg:search}, the source sentence cannot be
completely covered, i.e. there are still some ``leftover'' words. 
The three examples show that the leftover words in both sentences,
which could not be aligned (from $s_1$ to $s_2$ and from $s_2$ to $s_1$). 

The first failure is due to our checker's inability to resolve a
co-reference between ``he'' and ``De Sole''. 
The second failure is due to the mismatch between
``around the world'' and ``Americans''. However, both GPT3.5 and the 
label consider the two sentences semantically equivalent.
The third example shows many more leftover words. Still both 
GPT3.5 and the 
label consider the two sentences semantically equivalent.
These examples clearly indicate there is room for future optimization of our
``Yes'' prover. 

\begin{table*}[ht]
	\centering
	\caption{An example of phrase alignments between $s_1$ and $s_2$, 
		which is a ``Yes'' prover accepted cases. \label{tab:AppB_accepted}}
	\begin{tabular}{rl}
		\toprule
		\multicolumn{1}{l}{$s_1$} & $s_2$ \\
		\midrule
		\parbox{0.45\textwidth}{Sheena Young of Child, the national infertility support network, hoped the guidelines would lead to a more "fair and equitable" service for infertility sufferers.} &
		\parbox{0.45\textwidth}{Sheena Young, a spokesman for Child, the national infertility support network, said the proposed guidelines should lead to a more "fair and equitable" service for infertility sufferers.} \\ \\ 
		\multicolumn{2}{c}{ \bf Phrases from $s_1$ (left) aligned with phrases from $s_2$ (right)} \\[10pt] 
		\parbox{0.28\textwidth}{``Sheena Young of Child, the national infertility support network,''} &
		\parbox{0.32\textwidth}{``Sheena Young, a spokesman for Child, 
			the national infertility support network,''} \\[12pt]
		\parbox{0.35\textwidth}{``the national infertility support network,''} 
		&
		\parbox{0.35\textwidth}{``the national infertility support network,''} \\[10pt] 
		\parbox{0.30\textwidth}{lead to a more "fair and equitable" service for infertility sufferers,} &
		\parbox{0.30\textwidth}{lead to a more "fair and equitable" service for infertility sufferers,} \\[12pt]
		\parbox{0.36\textwidth}{the guidelines would lead to a more "fair and equitable" service for infertility sufferers,} &
		\parbox{0.40\textwidth}{the proposed guidelines should lead to a more "fair and equitable" service for infertility sufferers,} \\[10pt]
		\parbox{0.15\textwidth}{	Sheena Young,} &
		\parbox{0.15\textwidth}{Sheena Young,} \\[10pt]
		\parbox{0.4\textwidth}{hoped the guidelines would lead to a more "fair and equitable" service for infertility sufferers,} &
		\parbox{0.4\textwidth}{said the proposed guidelines should lead to a more "fair and equitable" service for infertility sufferers,} \\[10pt]
		\midrule
	\end{tabular}
\end{table*}

\subsubsection{An example accepted by the checker}

Table~\ref{tab:AppB_accepted} shows an example of ``Yes'' case accepted
by our checker. The proof is shown as alignments between the phrases 
in the two sentences. Note that only phrases corresponding to a
sub-tree root of a matching path ($matchingpath()$ in Algorithm~\ref{alg:search})
are shown. The compositional nature of the matching is that
for every matching pair, there exists at least one
pair of matching sub-phrases contained in the two phrases.
(This is by definition, according to our $matchingpath()$ design).

\subsubsection{The five GPT3.5-generated paraphrases rejected by itself}
\label{fiveNOcases}

Table~\ref{CheckervsGPT} earlier shows that GPT3.5 considered five of its
own generated paraphrases as semantic non-equivalence. 
Table~\ref{tab:AppB_GPTgen} shows detail of these five paraphrases.  

According to our oracle-checker scheme, these five decisions 
would not go through the ``Yes'' prover. Instead, the ``No'' decisions would be
tested by our non-equivalence test. Nevertheless, Table~\ref{tab:AppB_GPTgen} 
shows the result of the ``Yes'' prover. 
It should be noted that when we ran the non-equivalence test on them,
all of them were rejected (under $\phi_{and}$ or $\phi_{or}$),
indicating that these decisions were not trusted by the checker. 

Table~\ref{tab:AppB_GPTgen}, on the other hand, shows that
the ``Yes'' prover could find proof for three of the five. 
Take the second case as an example. The phrase ``pull the pig from
the trough'' appeared in a news article about a lawsuit against
a former NYSE boss. The phrase referred to the boss being
forced out of the job and had nothing to do with an actual
farmyard situation. One can argue either way if the sentence
and its paraphrase should be considered as semantically equivalent. 

Because our non-equivalence test rejects all five cases, we can say
that those ``No'' decisions are not trustable. 
However, we {\it cannot} say that just because our ``Yes'' prover accepts
the first three and their ``No'' decisions are not trustable, 
those three should be considered as semantically
equivalent. It is important to note that our ``Yes'' prover is
{\it not} designed to make this type of claim. 
Our ``Yes'' prover is only involved with a ``Yes'' decision. 
If our intent is to run a ``Yes'' prover {\it after} a ``No''
decision is rejected, we need to design a new prover. 
This might be a possible future work.  

The five examples further illustrate the subtlety for
deciding on semantic equivalence. It is important that one does
not look at them from a correct/incorrect point of view.
Optimization for this absolution of correct vs. incorrect 
is {\it never} the objective of the work. 
Instead, we look at semantic equivalence from an
accept/reject point of view. In our scheme,
this subjective view is provided by the oracle and checker
{\it as a pair}. It is not based on either oracle
or checker alone, as pointed out at the beginning of
section~\ref{sec02} before. 

\subsection{Examples from the ``No'' cases}
\label{app02.4}

Table~\ref{Trust} earlier shows the acceptance rates on those ``No'' cases. 
Table~\ref{tab:AppB_NoReject} below provides examples from those rejected
``No'' cases. For each rejected ``No'' decision, our non-equivalence test
provides a {\it triangular evidence} to demonstrate why the decision
by GPT3.5 is rejected. 

Without loss of generality, a {\it triangular evidence} is in the 
form of a 3-tuple $(s_1, s_2, p)$ where $p$ is a semantically
equivalent paraphrase of $s_1$. 
GPT3.5 says that $s_1$ and $s_2$ are not semantically equivalent,
i.e. this is how we got the ``No'' decision to begin the test. 
In the test, there can be two outcomes.
The desired outcome is that the test finds an indifferentiable
$p$ (see Figure~\ref{nonisomorphism}) where $p$ is aligned with $s_2$. 
Then, the checker asks GPT3.5 whether or not $s_2$ and $p$ are
semantically equivalent. The key for a triangular evidence
is that GPT3.5 says $s_2$ and $p$ {\it are}
semantically equivalent. This is considered as a contradiction
to the earlier claim that $s_1$ and $s_2$ are not
semantically equivalent and further $p$ is a semantically
equivalent paraphrase of $s_1$.

The three sentences in each example in Table~\ref{tab:AppB_NoReject} 
form a triangular evidence to justify the rejection of the 
respective ``No'' decision. 
The difference between the first three examples and the last
three examples is in whether or not an indifferentiable $p$
could be found and was used in the test. 

\balance 

In the last three examples, no indifferentiable
paraphrase could be found. As a result, the test randomly choose
a paraphrase from a set of paraphrases. 
As explained with Table~\ref{Trust} before (e.g. under the $\bar{A}\bar{I}$
category), a randomly-chosen paraphrase is less effective for
rejecting a ``No'' decision. But in these three cases, the test
did find a triangular evidence with a random
paraphrase and hence, rejected the ``No'' decision.

Once again, under our oracle-checker scheme a rejected ``No''
decision does not mean that the answer should be a ``Yes''.
Our scheme can only lead to two usable outcomes: 
an accepted ``Yes'' decision and an accepted ``No'' decision.
Any decision that is rejected, only means that the
decision is not trusted by the checker. 
These outcomes are a consequence of our starting 
assumption that there is no universally-agreeable
definition of ``semantic equivalence''.
As a result, we rely on interactions between
an oracle and a checker to decide acceptance of a 
semantic equivalence decision, and
when this acceptance does take place, the subjective view
of the checker is incorporated.

        % Wim Wenders directed a widely praised film of the same name, based on the sessions. & 
        % A widely praised film of the same name was directed by Wim Wenders. &  
        % No & Yes \\
        % \multicolumn{4}{p\textwidth}{\it {\bf GPT3.5 Explanation:} The two sentences are semantically equivalent as they both convey the same information that Wim Wenders directed a widely praised film of the same name.} \\ \midrule

\begin{table*}[ht]
    \centering
    \caption{Five sentence pairs where GPT3.5 decided that its own generated paraphrase is not semantically equivalent to the original sentence. Our ``Yes'' prover accepted three and rejected two. The leftover words after alignment are shown for the two reject cases. \label{tab:AppB_GPTgen}}
    \begin{tabular}{p{0.4\textwidth}p{0.4\textwidth}c}
    \toprule
         s1 & GPT3.5 generated & ``Yes'' prover \\         
    \midrule
    % 1
         Marisa Baldeo stated, however, the authority's official uniform policy says "they are not supposed to wear anything on their heads but a NYC transit depot logo cap. &  
         Marisa Baldeo mentioned that the official uniform policy of the authority specifies that employees should only wear a NYC transit depot logo cap on their heads and nothing else. & 
         Accept \\
         \multicolumn{3}{p\textwidth}{\it {\bf GPT3.5 Explanation:} The two sentences are not semantically equivalent. The first sentence mentions that employees are not supposed to wear anything on their heads except for a specific cap, while the second sentence states that employees should only wear the specified cap on their heads and nothing else. The use of 'not supposed to wear anything' in the first sentence and 'should only wear' in the second sentence creates a difference in meaning.}
         \\ \midrule 
    % 2
        Sean Harrigan of Calpers, the California fund, said: "Today we are trying to pull the pig from the trough. &  
        Sean Harrigan from Calpers, the California fund, stated: 'Our current goal is to remove the pig from the feeding area.' & 
        Accept \\
        \multicolumn{3}{p\textwidth}{\it {\bf GPT3.5 Explanation:} The two sentences are not semantically equivalent. The first sentence implies a negative connotation, suggesting misuse of funds, while the second sentence has a more neutral tone, focusing on a goal of removing the pig from the feeding area.}
        \\ \midrule 
    % 3 
        Tenet has been under scrutiny since November, when former Chief Executive Jeffrey Barbakow said the company used aggressive pricing to trigger higher payments for the sickest Medicare patients. &  
        Tenet has been under investigation since November, following former CEO Jeffrey Barbakow's statement that the company employed aggressive pricing to elicit increased payments for the most ill Medicare patients. & 
        Accept \\
        \multicolumn{3}{p\textwidth}{\it {\bf GPT3.5 Explanation:} The two sentences convey similar information about Tenet being under scrutiny or investigation since November due to the company's use of aggressive pricing to increase payments for the sickest Medicare patients. However, the wording and structure of the sentences are different, and there are some variations in the specific phrasing and details, so they are not considered paraphrases.} 
        \\ \midrule
    % 4 
        Another was in serious condition at Northwest Medical Center in Springdale. & 
        One more person was in critical condition at Northwest Medical Center in Springdale. & 
        Reject \\ 
        \multicolumn{3}{p\textwidth}{\it {\bf Post-alignment leftover: } $s_1$: \{ \textcolor{orange}{Another} \}, GPT3.5 generated: \{ \textcolor{blue}{One more person} \} } \\
        \multicolumn{3}{p\textwidth}{\it {\bf GPT3.5 Explanation:} The two sentences are not semantically equivalent. The first sentence states 'serious condition' while the second sentence states 'critical condition', which are not synonymous.} 
        \\ \midrule
    % 5
        Meanwhile, rival contender, General Electric's NBC, submitted a letter of interest, a source familiar with the matter said. & 
        At the same time, General Electric's NBC, a competing contender, sent a letter of interest, according to a source who is knowledgeable about the situation. &
        Reject \\
        \multicolumn{3}{p\textwidth}{\it {\bf Post-alignment leftover: } $s_1$: \{ \textcolor{orange}{Meanwhile} \}, GPT3.5 generated: \{ \textcolor{blue}{At the same time} \} } \\
        \multicolumn{3}{p\textwidth}{\it {\bf GPT3.5 Explanation:} The two sentences have similar meanings but are not exact paraphrases. The first sentence uses 'meanwhile' to indicate a temporal relationship, while the second sentence uses 'at the same time.' Additionally, the first sentence refers to 'rival contender' while the second sentence uses 'competing contender.' The choice of words and sentence structure differ slightly, making them not exact paraphrases.} \\
        \bottomrule
    \end{tabular}
\end{table*}

\begin{table*}[ht]
	\centering
	\caption{Examples of ``triangular evidence'' for rejecting ``No'' cases.
		For each example, three sentences are shown where the GPT3.5 says
		the 1st and the 2nd sentences are not semantically equivalent 
		but both are equivalent to a paraphrase of the 1st sentence. 
		For the first three cases, the paraphrase $p$ is indifferentiable
		(see section~\ref{sec04.3} for the definition of being indifferentiable). 
		For the remaining three cases, the paraphrase is not indifferentiable,
		and thus is randomly chosen among 10 generated paraphrases. \label{tab:AppB_NoReject}}
	\begin{tabular}{p{0.31\textwidth}p{0.31\textwidth}p{0.31\textwidth}}
		\toprule
		1st sentence & 2nd sentence, where GPT3.5 says that
		it is {\it not} semantically equivalent to the 1st sentence. & GPT3.5's paraphrase of the 1st sentence, 
					is semantically equivalent to the 2nd sentence, according to itself. \\
		\midrule 
		\multicolumn{3}{c}{\bf Cases where the paraphrase $p$ of the 1st sentence is indifferentiable:} \\[5pt]
		Police warned residents on Friday not to travel alone and to avoid convenience stores and service stations stores at night.&
		Police advised residents Friday not to travel alone to convenience stores and to be watchful.&
		Police on Friday cautioned residents against traveling alone and recommended avoiding convenience stores and service stations at night. \\[5pt]
		The company has expanded those basic services, offering payment and even financing. &
		The company has expanded into providing other services for buyers, including payment services. &
		The company has broadened its range of fundamental services, providing payment and even financing. \\[10pt]
		Its attackers had to launch their strike from hundreds of metres away, with a makeshift rocket launcher disguised as a portable electric generator. &
		Its attackers had to fire their rockets from hundreds of yards away, with a makeshift launcher hidden in a portable electric generator. &
		The assailants had to initiate their attack from a distance of several hundred meters using an improvised rocket launcher that was camouflaged as a portable power generator. \\ [20pt]
		\midrule 
		\multicolumn{3}{c}{\bf Cases where the paraphrase is randomly selected, because no indifferentiable paraphrase $p$ could be found:} \\[5pt]
		Rudder was most recently senior vice president for the Developer \& Platform Evangelism Business. &
		Senior Vice President Eric Rudder, formerly head of the Developer and Platform Evangelism unit, will lead the new entity. &
		Rudder's most recent role was as the senior vice president for the Developer \& Platform Evangelism Business. \\[20pt]
		"I think you'll see a lot of job growth in the next two years," he said, adding the growth could replace jobs lost. &
		"I think you'll see a lot of job growth in the next two years," said Mankiw.&
		According to him, there is likely to be a substantial increase in employment opportunities in the next two years, with the potential for the growth to compensate for lost jobs. \\[10pt]
		No dates have been set for the criminal or civil cases, but Shanley has pleaded not guilty.&
		No dates have been set for the civil or the criminal trial. &
		No specific dates have been arranged for the criminal or civil cases, despite Shanley's plea of not guilty. \\ [10pt]
		\bottomrule
	\end{tabular}
\end{table*}

%%%%%%%%%%%%%%%%%%%%%%%%%%%%%%%%%%%%%%%%%%%%%%%%%%%%%%%%%%%%%%%%%%%%%%%%%%%%%%%

\end{document}